\documentclass[notoc]{kling-keye}
\def\VNarxiv{1}
\usepackage{xspace,adjustbox,flafter}
\usepackage[shortlabels,inline]{enumitem}
\usepackage{multirow}
\usepackage{makecell}
\usepackage{tikz}
\usetikzlibrary{calc}
\usetikzlibrary{positioning,arrows.meta,fit,backgrounds,calc,shapes.geometric}
\usepackage{pgfplots}
\pgfplotsset{compat=1.17}
\usepackage{fontawesome5}   
\usetikzlibrary{shadows,fadings,patterns,decorations.pathreplacing,shapes.misc}
\definecolor{cOurs}{HTML}{E8604C}     
\definecolor{cOursInk}{HTML}{B8432F}
\definecolor{cSeek}{HTML}{3D78C9}     
\definecolor{cSeekInk}{HTML}{2A5A9E}
\definecolor{cTeal}{HTML}{22A08C}     
\definecolor{cTealInk}{HTML}{167565}
\definecolor{cAmber}{HTML}{D9A21B}    
\definecolor{cAmberInk}{HTML}{8F6A08}
\definecolor{cPlan}{HTML}{3D78C9}    
\definecolor{cPlanInk}{HTML}{2A5A9E}
\definecolor{cSlate}{HTML}{7D8597}    
\definecolor{cInk}{HTML}{141413}
\definecolor{cInkB}{HTML}{52514E}
\definecolor{cInkC}{HTML}{85837C}
\definecolor{cGrid}{HTML}{E4E3DF}
\tikzset{
  vn panel/.style={rounded corners=3pt, line width=0.6pt},
  vn chip/.style={rounded corners=2pt, inner sep=1.6pt, font=\sffamily\bfseries\scriptsize,
    text=white},
  vn card/.style={rounded corners=2.5pt, fill=white, line width=0.6pt,
    drop shadow={opacity=.12, shadow xshift=.5pt, shadow yshift=-.5pt}},
  vn call/.style={-{Stealth[length=4pt,width=3.4pt]}, line width=0.9pt, draw=cPlan},
  vn return/.style={-{Stealth[length=4pt,width=3.4pt]}, line width=0.9pt, draw=cOurs},
}
\usepackage{pifont}

\newcommand{\dev}[1]{#1}

\newcommand{\method}{VideoNexus\xspace}
\newcommand{\planner}{Planner\xspace}
\newcommand{\finding}{Finding\xspace}
\newcommand{\cmark}{\ding{51}}
\newcommand{\xmark}{\ding{55}}
\newcommand{\supp}{the supplementary material\xspace}

\newcommand{\maketitlesupplementary}{\begin{center}\Large\sffamily\bfseries\color{KeyePrimary}Appendix\end{center}}

\newcommand{\vnauthor}[2]{\begin{minipage}[t]{0.235\textwidth}\centering\sffamily\normalsize\textbf{#1}\\[3pt]{\normalfont\fontsize{8}{10}\selectfont #2}\end{minipage}}
\hypersetup{pdftitle={VideoNexus: Video Perception Beyond the Reasoning Context},pdfauthor={Yancheng Long, Xuanyu Zheng, Wei Chen, Luoyuan Zhang, Haoyang Xu, Bin Wen, Tingting Gao}}

\providecommand{\eg}{\emph{e.g.}\xspace}

\title{VideoNexus: Video Perception Beyond\\the Reasoning Context}
\author{%
\vnauthor{Yancheng Long}{The Chinese University\\of Hong Kong}\hfill
\vnauthor{Xuanyu Zheng\textsuperscript{\ensuremath{\ddagger,\dagger}}}{KlingAI Research}\hfill
\vnauthor{Wei Chen}{The Hong Kong University\\of Science and Technology}\hfill
\vnauthor{Luoyuan Zhang}{KlingAI Research}\\[12pt]
\vnauthor{Haoyang Xu}{KlingAI Research}\hspace{12pt}
\vnauthor{Bin Wen\textsuperscript{\ensuremath{\dagger}}}{KlingAI Research}\hspace{12pt}
\vnauthor{Tingting Gao}{KlingAI Research}}
\affiliation{\textsuperscript{\ensuremath{\ddagger}}\,Project Lead\qquad\textsuperscript{\ensuremath{\dagger}}\,Corresponding authors}
\projectlinks{}
\headertext{VideoNexus: Video Perception Beyond the Reasoning Context}
\reportdate{November 16, 2026}
\reportabstract{Understanding long videos requires locating evidence, resolving visual details, and integrating observations into an answer.
Even when relevant evidence is present, performing these operations within long, dense visual sequences burdens both perception and reasoning.
Video agents enable selective inspection, but returning frames to the planner still accumulates visual information in its reasoning context.
Delegating perception reduces this burden, yet sparse sampling can miss decisive details, and per-frame descriptions can preserve context growth.
We introduce \method, a training-free framework that separates detailed video perception from evidence-driven planning and reasoning.
The planner controls where to inspect, what to resolve, and how densely to sample.
Perception sub-agents process native audio--video in independent contexts and return compact, temporally grounded evidence for iterative search and verification.
Using Gemini~3.8 Flash for both planning and perception, \method achieves 88.6\% accuracy on full LVBench, exceeding same-backbone static ingestion and built-in agentic processing by 3.6 and 1.1 points, respectively, at 44\% of the latter's measured API cost.
A paired LVBench subset ablation shows lower planner input and higher accuracy.
With Kimi~K3 as the shared backbone, \method improves over the official Video-GAIA ReAct harness by 8.5 points in our evaluation.
These results support compact evidence exchange for combining detailed observation with efficient planning and reasoning.
}
\begin{document}
\maketitle
\reportcontents
\section{Introduction}
\label{sec:intro}

\begin{figure*}[t]
  \centering
  \begin{adjustbox}{max width=\linewidth}
\input{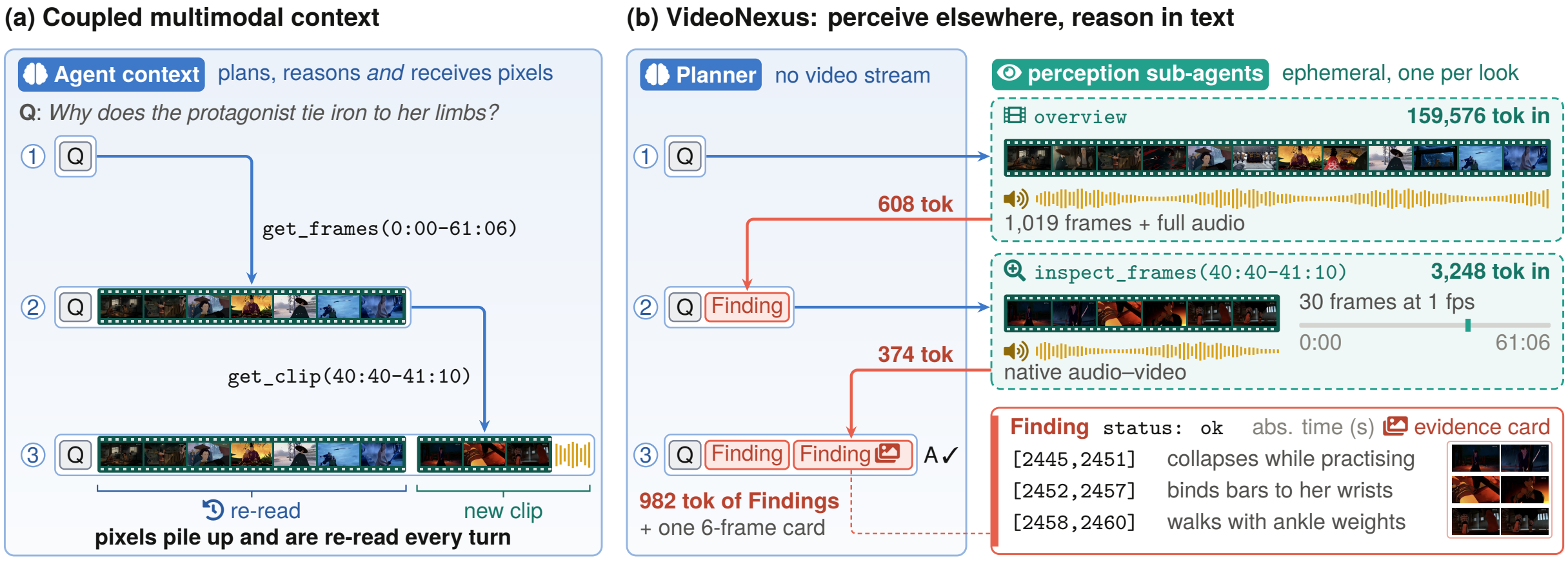}
\end{adjustbox}
  \caption{\textbf{Coupled versus decoupled multimodal context.} Each row shows the reasoning context at one turn. \textbf{(a)}~Retrieved video and audio accumulate alongside reasoning. \textbf{(b)}~\method returns compact, temporally grounded \finding{}s and a small evidence card from independent perception calls. Colors distinguish video (sea green), audio (gold), reasoning (blue), and findings (coral).}
  \label{fig:teaser}
\end{figure*}

Understanding a long video involves finding the evidence that matters to a question, examining its visual details, and reasoning about how observations fit together.
These operations are closely related: an initial observation may suggest a location to revisit, a detail to verify, or a new question to resolve.
Even when all relevant evidence is included in the input, a model must locate and interpret it among substantial amounts of unrelated visual information.
A one-hour video sampled at one frame per second produces hundreds of thousands of visual tokens~\cite{wang2024lvbench}.
\emph{Static ingestion} places evidence localization, perception and reasoning within this shared multimodal context~\cite{geminiteam2024gemini15,bai2025qwen25vl}; denser input provides more detail but also increases the information the model must process together.

\emph{Video agents} allow this process to unfold through interaction: the model chooses where to look, evaluates the returned evidence, and decides whether another observation is needed~\cite{wang2024videoagent,fan2024videoagent,wang2025videotree,zhang2025dvd}.
EVA~\cite{zhang2026eva} learns query-driven observation policies, and VideoTemp-o3~\cite{liu2026videotempo3} jointly learns temporal grounding and question answering with localization refinement. Alongside these advances in selecting observations, the interface that delivers them to the reasoning context remains consequential.
When tools return raw frames or clips directly to the planner, visual inputs accumulate in the context that also guides subsequent actions and integrates evidence (\cref{fig:teaser}a).
Adaptive selection changes which media is read, but the planner continues to process that media alongside its reasoning history.
In our controlled comparison, this coupled design uses 1.47$\times$ as much planner input as returning evidence from independent perception calls (\cref{sec:exp:coupling}).

Delegating perception to separate models addresses this context burden and is already established in video agents.
VideoSeek~\cite{lin2026videoseek} and VideoXAgent~\cite{yang2026videoxagent}, for example, allow a planner to act on textual observations.
Two issues remain.
First, sparse sampling can omit brief events or fine details needed to verify an answer, motivating inspection at a density appropriate to the local task.
Second, a text interface can preserve the dependence between observation size and sampling density.
VideoSeek's global scan returns one description per frame, averaging 28 returned tokens per frame read in our re-run (\cref{sec:exp:bounded}).
Increasing its scan budget would therefore also lengthen the report retained for reasoning.
Separating model calls alone does not determine either how closely the agent can inspect a moment or how much information it must carry forward.

We study how to give video agents access to detailed observations while controlling what each observation adds to the reasoning context.
The required detail varies across the process: broad exploration locates candidate moments, while local verification may require denser sampling around a specific event.
This motivates treating each perception call as a focused evidence-gathering task, with its location and sampling density chosen by the planner and its output organized around the question being resolved.

We introduce \method, a training-free framework for video perception beyond the reasoning context (\cref{fig:teaser}b).
The \planner formulates local tasks, selects intervals and sampling rates, integrates returned evidence, and decides when to stop.
Perception sub-agents examine native audio--video in independent, short-lived contexts.
Overview returns a compact set of candidate intervals for navigation; Inspect returns a \finding with claims anchored to the absolute source timeline and explicit evidence or uncertainty status.
Both global and local reports limit the information returned instead of describing every sampled frame.
In the final configuration, small local evidence cards accompany visual findings for verification.
The planner can thus request further detail without retaining the complete perceptual input from each call.
\method runs on a stock agent runtime with Model Context Protocol tools and requires no training.

Our experiments examine context use, observation size and task performance.
With Gemini~3.8 Flash as the shared backbone, a paired LVBench ablation reduces planner input by 32\% while increasing accuracy from 86.5\% to 89.0\% in that run.
On full LVBench~\cite{wang2024lvbench}, \method achieves 88.6\%, compared with 85.0\% for static ingestion and 87.5\% for Gemini's agentic mode, at 44\% of the latter's measured API cost.
It reads 621 frames per question on average and outperforms a same-backbone VideoSeek re-run by 7.2 points at lower cost; the comparison also differs in audio access, analyzed in \supp.
With Kimi~K3 as the shared backbone on Video-GAIA~\cite{zhang2026videogaia}, it improves over the official ReAct harness by 8.5 points in our evaluation; further experiments examine transfer across planner models.

Our contributions are:
\begin{itemize}[leftmargin=*,itemsep=1pt,topsep=2pt]
  \item We analyze the coupling between video observation and reasoning context, including how per-frame text can preserve this dependence after perception is delegated.
  \item We introduce \method, combining planner-directed perception with bounded, temporally grounded evidence exchange for global exploration and local verification.
  \item We evaluate the design through controlled ablations and per-call measurements, together with same-backbone accuracy--cost comparisons and transfer experiments.
\end{itemize}

\section{Related Work}
\label{sec:related}

\paragraph{Long contexts and video representations.}
Frontier MLLMs accept long native videos~\cite{geminiteam2024gemini15,comanici2025gemini25,bai2025qwen25vl,keye2026vl2}; other work extends context~\cite{zhang2024longva} or compresses visual tokens~\cite{shen2024longvu}. Long-video benchmarks expose difficulties in retrieving and reasoning over detailed evidence~\cite{fu2025videomme,wu2024longvideobench,wang2024lvbench}, and required sampling density varies across questions~\cite{qu2025lsdbench}. Text-only studies likewise show that access to a long context does not ensure effective use of its contents~\cite{liu2024lost}.
Real-time systems support full-duplex audiovisual interaction~\cite{cui2026minicpmo45} or parallel listening, writing and speaking~\cite{zhang2026lws}, addressing interaction constraints distinct from long-video evidence search.
Language-mediated representations~\cite{zeng2023socratic,zhang2024llovi} and searchable video memories~\cite{luo2024videorag,wang2024videoagent,fan2024videoagent,zhang2025dvd,cai2026cfd} organize evidence for later reasoning. Several methods combine initial representations with query-conditioned retrieval, refinement or inspection~\cite{ma2025drvideo,wang2025videotree,yin2026haven,liu2026videotapestry}; they are not uniformly fixed, question-independent indices. Sequential memory construction~\cite{chen2026memdreamer,chen2026sprout} and compositional visual programs~\cite{suris2023vipergpt,gupta2023visprog,min2024morevqa} offer further ways to structure perception.
Precomputed representations can amortize work across questions, while query-time inspection can revisit details omitted from the initial representation. We focus on the interface between such inspection and the planner's reasoning context.

\paragraph{Learned policies and working memory.}
OmniAgent~\cite{xing2026omniagent} learns active perception; EVA~\cite{zhang2026eva} trains a summary--plan--action--reflection loop; and VideoTemp-o3~\cite{liu2026videotempo3} jointly trains temporal grounding and question answering with localization refinement.
VideoSEAL~\cite{qiu2026videoseal} separates planning from final answer authority for visual verification, whereas VideoLoop~\cite{xu2026videoloop} provides plug-and-play working-memory rewriting. These address different parts of an agent's evidence-processing pipeline.
In adjacent settings, Thyme~\cite{zhang2025thyme} learns image-processing code for visual reasoning, SpatialReward~\cite{long2026spatialreward} uses explicit spatial reasoning to evaluate image edits, and DeltaVid~\cite{yang2026deltavid} improves fine-grained video perception through cross-video difference supervision.
\method studies a training-free observation interface. Learned policies, improved perception and compact evidence exchange are complementary; our experiments do not compare against these trained policies.

\paragraph{Query-time and delegated perception.}
Recent agents combine query-conditioned observation with tool orchestration. AVP~\cite{wang2025avp} separates coarse scanning from detailed inspection; VideoResearcher~\cite{ye2026videoresearcher} evolves tools through solving and tool-refinement loops; VideoXAgent~\cite{yang2026videoxagent} integrates a broad tool set; and VideoSeek~\cite{lin2026videoseek} navigates through overview, skim and focus operations.
Related approaches explore multi-agent coordination, observation routing and task-specific tools~\cite{yan2026symphony,wang2026route2look,gao2025avi,zhang2026vesta,tao2025omniagent,fan2026medclaw,liang2026vidharness}; GoP-Agent~\cite{chi2026gopagen} exploits compressed-video motion information. These systems differ in their memory and observation interfaces.
Delegated perception is established practice. Our focus is how returned reports scale with media inspected during global navigation and local verification. In our VideoSeek re-run, global scanning returns per-frame descriptions, while local focus already returns a concise answer. \method returns up to eight candidate intervals from Overview and task-specific \finding{}s from Inspect; \cref{sec:exp:bounded} measures their output growth.
LAVE~\cite{wang2026lave} instead injects latent visual evidence into reasoning. Our main configuration also retains limited visual evidence through small cards, while explicit uncertainty can prompt further inspection.

\paragraph{Commercial modes and evaluation.}
OpenAI's o3 reasons with images, and Gemini provides agentic video processing. The Qwen3.8-Omni report~\cite{qwen2026qwen38omni} also evaluates agentic video understanding. Open-world agents combine video reasoning with research~\cite{liu2026thinkingbeyondvideos}; VideoDR~\cite{liu2026videodr} and Video-GAIA~\cite{zhang2026videogaia} provide benchmarks with ReAct-style baselines~\cite{yao2023react}.
Static ingestion remains competitive~\cite{rege2026egagent}, motivating our same-backbone static comparison. TRACE~\cite{liu2026trace} evaluates visual evidence coverage. REPAIR~\cite{alhamidi2026repair} highlights variation across repeated agent executions: our full-set paired tests address within-run differences, not repeat-run reproducibility.
Anthropic's multi-agent research system also delegates research to sub-agents that return condensed results. \method examines this design for video, where the detail of each observation and the size of its report can be controlled separately.

\section{Method}
\label{sec:method}

\begin{figure*}[!t]
  \centering
  \begin{adjustbox}{max width=\linewidth}
\input{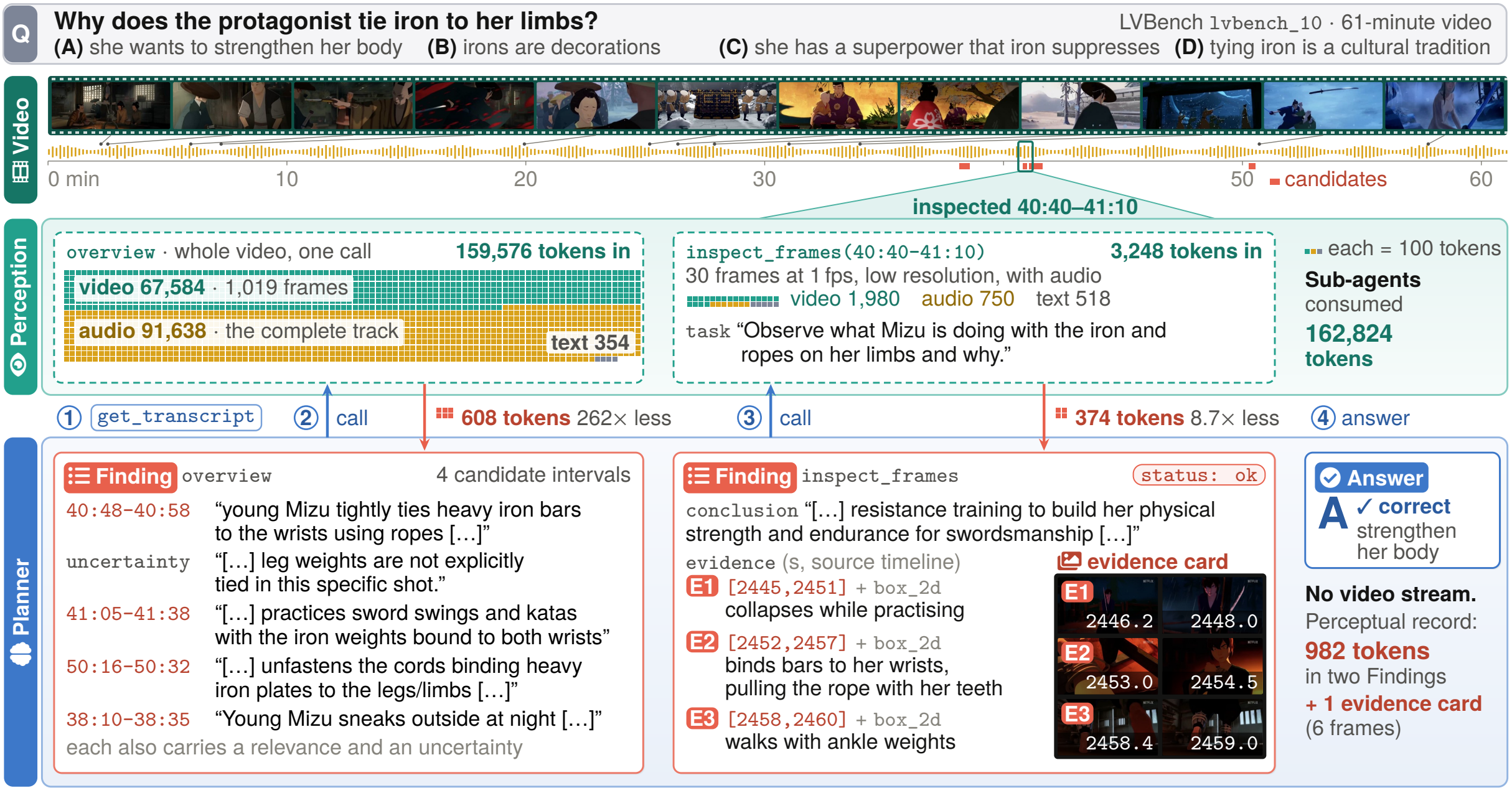}
\end{adjustbox}
  \caption{\textbf{\method on one recorded LVBench question.} Top: the 61-minute source video, candidate intervals (coral), and the selected 30-second window. Middle: independent perception contexts, with each cell representing 100 tokens. Bottom: an Overview report, an Inspect \finding, and a six-frame evidence card on the source timeline. Perception consumes 162,824 tokens; the planner receives 982 finding tokens and one card.}
  \label{fig:trajectory}
\end{figure*}

\subsection{Coupled and decoupled multimodal contexts}
\label{sec:method:formal}

Given a video $V$ of duration $T$ and a question $q$, an agent alternates between reasoning and tool calls until it answers.
Let $C_t$ be the context of the model that plans at step $t$, $a_t$ its action and $o_t$ the observation returned to it, so that $C_{t+1}=C_t\oplus a_t\oplus o_t$.
In a \emph{coupled} agent the observation is the media itself, $o_t=V[s_t,e_t]$ sampled at frame rate $f_t$, of size
\begin{equation}
  |o_t| \approx (e_t-s_t)\,\big(f_t\,\tau_v + \tau_a\big),
  \label{eq:coupled}
\end{equation}
where $\tau_v$ is the token cost of one frame and $\tau_a$ that of one second of audio; for Gemini at low media resolution $\tau_v{=}66$ and $\tau_a{\approx}25$.
In a history-preserving implementation, these tokens stay in $C_t$ and are attended to at later steps; truncation or memory compression can change this growth.
Replacing pixels by per-frame descriptions of $w$ tokens~\cite{wang2024videoagent,lin2026videoseek} changes the constant but not the scaling, $|o_t|\approx (e_t-s_t)\,f_t\,w$: without further compression or selection, increasing sampling density also increases the observation retained by the planner.
In a \emph{decoupled} agent, perception is a separate function $\Phi$ that runs in its own context and returns a compact description,
\begin{equation}
  o_t = \Phi\big(V[s_t,e_t],\, f_t,\, \kappa_t\big), \qquad |o_t| \ll (e_t-s_t)\,(f_t\tau_v+\tau_a),
  \label{eq:decoupled}
\end{equation}
where $\kappa_t$ is a local task written by the planner.
The intended effect is to make context growth depend primarily on the evidence needed for the local task, allowing $f_t$ to increase without retaining every frame or its description.
The planner integrates evidence and selects actions, while $\Phi$ performs task-specific perceptual reasoning. The roles can share a backbone or use different models.

\subsection{\method}
\label{sec:method:agents}

\method consists of a \planner that never receives the video stream and a set of perception sub-agents exposed as tools (\cref{fig:trajectory}; a schematic and the full tool list are in \supp).
The \planner receives the question and a short policy (a ``skill'' of about ten sentences) that imposes no fixed call order and states that transcripts may contain errors, that Overview returns navigation clues rather than evidence, that visual claims require visual evidence, and to stop once the answer is supported.
Every perception call runs an independent model call on native media, in a context that is discarded afterwards.

\textbf{Transcribe} runs ASR once per video, independently of the question, and returns segments with absolute times; it is cached and shared across questions.
\textbf{Overview} performs a query-conditioned, globally sparse scan of the native video (at most 1\,fps and 1,024 frames, at low resolution, with the full audio track) and returns a short summary and up to eight candidate intervals, each with an observation, a relevance estimate and an uncertainty note.
Its prompt states that sparse sampling cannot establish absence and forbids choosing an answer option, so Overview is a \emph{locator}, not a judge; it is limited to two calls per question.
\textbf{Inspect}, the main evidence source, reads an interval $[s,e]$ at a frame rate chosen by the \planner, with an open-ended local task such as ``describe what the person places on the red armchair and when''.
The skill asks the \planner to phrase local tasks neutrally, and the perception prompt forbids selecting an option of the parent question, so a \finding reports what the interval shows rather than echoing the \planner's hypothesis.
For open-world questions the same pattern extends to source-resolution crops and to web search and page extraction~\cite{zhang2026videogaia}; the LVBench runs use only Transcribe, Overview and Inspect (with crops).

\subsection{The \finding protocol}
\label{sec:method:finding}

Each Inspect call returns a JSON \finding with a \texttt{status} $\in$ \{\texttt{ok}, \texttt{absent}, \texttt{uncertain}, \texttt{failed}\}, where \texttt{absent} refers only to the sampled observation; a one-sentence \texttt{conclusion}; what was \texttt{observed}; and \texttt{evidence} items $\{\textit{claim}, [t_0,t_1], \textit{modality}\}$ with modality $\in$ \{visual, audio, audio\_visual\} and times on the \emph{absolute} source timeline.
The harness validates every \finding: evidence whose time range falls outside the inspected interval is dropped, and an \texttt{ok} status without remaining evidence is downgraded to \texttt{uncertain}.
This validates structure and temporal range, not factual correctness. Overview caps candidates at eight; Inspect has no eight-item cap. Both have a generation budget independent of frame count. Bounded evidence exchange refers to this budget and compact report design; \cref{sec:exp:bounded} measures actual output growth.

The final configuration returns each visual \finding with an \emph{evidence card}: a single contact sheet containing at most six source frames selected from its cited evidence intervals, each labeled with its absolute time. Dense video and audio remain in the perception context.
The card grows with evidence, not with inspected duration; on full LVBench it amounts to 3.3 images and 2\% of the \planner's input tokens per question, with model-dependent effects on accuracy (\cref{sec:exp:coupling}).

\subsection{Native timeline addressing and implementation}
\label{sec:method:impl}
\label{sec:method:timeline}

Perception calls reference the original video file with start and end offsets and a frame rate, instead of cutting and re-encoding clips, and the perception model then reports times on the original timeline: in a probe with neutral prompts, all 12 responses returned absolute source times, while a physically cropped clip returned times relative to its own start (\supp).
\finding{}s from different calls, the transcript and the Overview can therefore be compared directly.
Each question runs in an isolated headless Claude Code session whose only built-in tool is the skill loader; perception tools are served over the Model Context Protocol, and a local bridge routes the \planner to any provider, replaying Gemini's thought signatures across turns.
Unless stated otherwise, the \planner and sub-agents share the model and high thinking effort, with at most 32 turns, 256 frames per Inspect and 10 local calls. Cumulative-budget implementation and observed usage are detailed in \supp.
The harness requires no training and no access to model weights.

\section{Experiments}
\label{sec:exp}

\subsection{Setup}
\label{sec:exp:setup}

\paragraph{Benchmarks.}
\textbf{LVBench}~\cite{wang2024lvbench} contains 1,549 four-way multiple-choice questions on 103 hour-scale videos in six categories: entity recognition (ER, 560), event understanding (EU, 526), key information retrieval (KIR, 196), reasoning (Rea, 147), summarization (Sum, 55) and temporal grounding (TG, 65).
We report accuracy on the full set and count invalid outputs as wrong.
Ablations use a fixed 200-question development subset stratified by category and covering all 103 videos (\emph{dev-200}); they are paired comparisons inside that subset, while the main Gemini~3.8 system comparison uses the full set, including these development questions.
\textbf{Video-GAIA}~\cite{zhang2026videogaia} contains 271 open-ended questions (L1: 62, L2: 209) whose answers require video evidence and web search; both arms are scored by one binary semantic judge (Kimi~K3) with the same prompt and reference answers (\supp).

\paragraph{Models and baselines.}
The main experiments use Gemini~3.8 Flash for both the \planner and perception, with high thinking effort and low media resolution; cross-model experiments use Gemini~3.0 Flash and Kimi~K3.
Hosted models are accessed through APIs; the reported backbone is shared across the main comparison.
\textbf{Static} feeds the whole native video with audio in one call at 1\,fps, capped at 2,048 frames.
\textbf{Gemini Agentic} is Google's server-side agentic media-processing mode with high thinking; it reports the media it navigates as image tokens, from which we estimate 464 frames per question (\supp).
\textbf{VideoSeek}~\cite{lin2026videoseek} is re-run from its official code on the same Gemini~3.8 Flash back-end at its published perception budget, with its own prompts, tools and budgets; we drive it through the provider's native interface with thought signatures replayed, without which over half of its steps return empty (\supp).
VideoSeek has no audio path, and LVBench ships no subtitles, so it never receives speech; we quantify this asymmetry in \supp.
On Video-GAIA we run the \textbf{official ReAct agent}, which keeps 20 uniformly sampled frames in context.

\paragraph{Cost.}
We report API cost per question at list prices (\$0.75 per million fresh input tokens, \$0.075 per million cached input tokens, \$3.75 per million output tokens including thinking), and \emph{cold-start} every question: caches shared across questions are repriced as fresh input, while caching within a question is kept (\supp). ASR cost is excluded for all methods.

\subsection{Main results on LVBench}
\label{sec:exp:main}

\begin{table*}[htbp]
  \caption{\textbf{LVBench results.} The main comparison (lower block) uses Gemini~3.8 Flash on all 1,549 questions, counting invalid outputs as wrong. The upper block reproduces GPT-5 results from VideoSeek~\cite{lin2026videoseek}: Static uses 384 uniformly sampled frames, and VideoSeek improves accuracy by 8.3/10.2 points without/with subtitles. Compare methods within each block. Raw media marks retrieved video/audio in the planner; frames/q counts all models. \method delegates its 621 video frames to perception and additionally returns 3.3 evidence-card images per question to the planner. Cost is cold-start API cost. $^\dagger$Estimated at 66 tokens/frame. --: unreported.}
  \label{tab:lvbench}
  \centering
  \small
  \setlength{\tabcolsep}{4.2pt}
  \maxsizebox{\textwidth}{!}{%
  \begin{tabular}{@{}llcrcccccccc@{}}
    \toprule
    Method & Backbone & \makecell{Raw media\\in planner?} & \makecell{Frames\\/ q} & Overall & ER & EU & KIR & Rea & Sum & TG & \makecell{Cost\\(\$/q)} \\
    \midrule
    \multicolumn{12}{@{}l}{\emph{Reported reference results: GPT-5}} \\
    Static (384 uniform frames) & GPT-5 & -- & 384 & 60.1 & -- & -- & -- & -- & -- & -- & -- \\
    \quad + subtitles & GPT-5 & -- & 384 & 66.5 & -- & -- & -- & -- & -- & -- & -- \\
    VideoSeek~\cite{lin2026videoseek} & GPT-5 & \xmark & 92.3 & 68.4 & -- & -- & -- & -- & -- & -- & -- \\
    \quad + subtitles & GPT-5 & \xmark & 27.2 & 76.7 & -- & -- & -- & -- & -- & -- & -- \\
    \midrule
    \multicolumn{12}{@{}l}{\textbf{Main comparison (our evaluations): Gemini 3.8 Flash}} \\
    Blind (question only, no media) & Gemini 3.8 Flash & -- & 0 & 40.5 & 39.6 & 40.7 & 39.8 & 41.5 & 49.1 & 38.5 & 0.014 \\
    Static (1\,fps, $\le$2048 frames) & Gemini 3.8 Flash & -- & 2{,}035 & 85.0 & 83.4 & 85.7 & 89.3 & 79.6 & 89.1 & 87.7 & 0.172 \\
    Gemini Agentic & Gemini 3.8 Flash & \cmark & 464$^\dagger$ & 87.5 & 86.2 & 88.0 & 90.3 & 82.3 & 92.7 & \textbf{92.3} & 0.319 \\
    VideoSeek~\cite{lin2026videoseek} (re-run) & Gemini 3.8 Flash & \xmark & 123 & 81.4 & 80.7 & 80.2 & 85.2 & 77.6 & 87.3 & 89.2 & 0.188 \\
    \midrule
    \textbf{\method} & Gemini 3.8 Flash & \xmark & 621 & \textbf{88.6} & \textbf{87.7} & \textbf{88.2} & \textbf{91.8} & \textbf{85.0} & \textbf{96.4} & 90.8 & 0.140 \\
    \bottomrule
  \end{tabular}}

\end{table*}

\begin{figure}[!t]
  \centering
  \begin{adjustbox}{max width=\linewidth}
\begin{tikzpicture}[
  f3guide/.style={draw=cOurs!55, line width=0.5pt, dash pattern=on 2pt off 1.5pt},
  f3lab/.style={font=\sffamily\scriptsize, inner sep=0.6pt},
  f3mk/.style={draw=white, line width=0.5pt, inner sep=0pt},
]
\pgfplotsset{f3base/.style={
  scale only axis, width=195pt, xmin=0, xmax=0.325, clip=false,
  axis lines*=left, axis line style={draw=cInkC, line width=0.4pt, -},
  xtick={0,0.1,0.2,0.3}, xticklabels={\$0,\$0.10,\$0.20,\$0.30},
  tick align=outside, major tick length=2pt, tick style={draw=cInkC, line width=0.4pt},
  tick label style={font=\sffamily\scriptsize, text=cInkB},
  label style={font=\sffamily\scriptsize, text=cInkB},
  major grid style={draw=cGrid, line width=0.4pt},
}}
\begin{axis}[f3base, name=pa, height=130pt, ymin=80, ymax=91,
  ytick={80,82,...,90}, yticklabels={80,82,84,86,88,90}, grid=major,
  xlabel={Cost per question (USD)}, xlabel style={yshift=1.5pt},
  ylabel={LVBench accuracy (\%)}, ylabel shift=-3pt, ylabel style={inner sep=1pt}]
\fill[cOurs!6] ([xshift=0.2pt]axis cs:0,88.57) rectangle (axis cs:0.1403,91);
\fill[pattern=north east lines, pattern color=cOurs!28] ([xshift=0.2pt]axis cs:0,88.57) rectangle (axis cs:0.1403,91);
\draw[cOurs!70, line width=0.5pt] ([xshift=0.2pt]axis cs:0,88.57) -- (axis cs:0.1403,88.57);
\node[f3lab, name=lQuad, fill=cOurs!6, rounded corners=1.5pt, inner sep=1.4pt, text=cOursInk,
  align=center, font=\sffamily\scriptsize\itshape] at (axis cs:0.0702,89.85)
  {nothing is cheaper\\and more accurate};
\draw[f3guide] (axis cs:0.1403,91) -- (axis cs:0.1403,80);
\coordinate (gA) at (axis cs:0.1403,80);
\draw[-{Stealth[length=3.5pt,width=3pt]}, cSlate, line width=0.6pt]
  (axis cs:0.014,82.3) -- (axis cs:0.014,80.15);
\node[f3lab, name=lBlind, fill=white, anchor=west, text=cInkB, align=left] at ([xshift=2.5pt]axis cs:0.014,81.2)
  {\textbf{Blind} control\\{\color{cInkC}40.5\% · \$0.014}};
\node[f3mk, name=mStatic, circle, fill=cTeal, minimum size=5.6pt] at (axis cs:0.1720,84.96) {};
\node[f3lab, name=lStatic, fill=white, anchor=north west, text=cTealInk, align=left] at ([xshift=3pt,yshift=-6.3pt]axis cs:0.1403,84.96)
  {\textbf{Static ingestion}\\{\color{cInkB}85.0\% · \$0.172}};
\node[f3mk, name=mGem, diamond, fill=cAmber, minimum size=7.2pt] at (axis cs:0.3191,87.48) {};
\node[f3lab, name=lGem, fill=white, anchor=north east, text=cAmberInk, align=right] at ([xshift=4pt,yshift=-2.5pt]mGem.south)
  {\textbf{Gemini Agentic}\\{\color{cInkB}87.5\% · \$0.319}};
\node[f3mk, name=mVS, regular polygon, regular polygon sides=3, fill=cSeek, minimum size=8pt] at (axis cs:0.1883,81.41) {};
\node[f3lab, name=lVS, fill=white, anchor=west, text=cSeekInk, align=left] at ([xshift=4pt]axis cs:0.1883,81.41)
  {\textbf{VideoSeek}\\{\color{cInkB}81.4\% · \$0.188}};
\node[name=mOurs, star, star points=5, star point ratio=2.3, minimum size=11pt, inner sep=0pt,
      fill=cOurs, draw=white, line width=0.5pt] at (axis cs:0.1403,88.57) {};
\node[f3lab, name=lOurs, fill=white, anchor=west, text=cOursInk, align=left] at ([xshift=6.5pt]axis cs:0.1403,88.57)
  {\textbf{\method{} (ours)}\\\textbf{88.6\%} · \textbf{\$0.140}};
\end{axis}
\begin{axis}[f3base, name=pb, at={(pa.south east)}, anchor=south west, xshift=64pt,
  height=130pt, ymin=0, ymax=80, axis y line=none,
  xlabel={Cost per question (USD)}, xlabel style={yshift=1.5pt}]
\draw[cGrid, line width=0.4pt] (axis cs:0.1,0) -- (axis cs:0.1,80) (axis cs:0.2,0) -- (axis cs:0.2,80)
  (axis cs:0.3,0) -- (axis cs:0.3,80);
\draw[f3guide] (axis cs:0.1403,0) -- (axis cs:0.1403,80);
\def\fthreebar#1#2#3#4{%
  \fill[#1!28] (axis cs:0,#2) rectangle (axis cs:#3,#2+10);
  \fill[#1]    (axis cs:#3,#2) rectangle (axis cs:#4,#2+10);
  \draw[white, line width=0.8pt] (axis cs:#3,#2) -- (axis cs:#3,#2+10);}
\fthreebar{cOurs}{58}{0.0938}{0.1403}
\fthreebar{cSeek}{33}{0.0901}{0.1883}
\fthreebar{cSeek}{8}{0.0421}{0.1403}
\node[f3lab, name=s0i, anchor=base, text=cOursInk] at (axis cs:0.0469,60.6) {0.094};
\node[f3lab, name=s0o, anchor=base, text=white]    at (axis cs:0.11705,60.6) {\textbf{0.046}};
\node[f3lab, name=s1i, anchor=base, text=cSeekInk] at (axis cs:0.04505,35.6) {0.090};
\node[f3lab, name=s1o, anchor=base, text=white]    at (axis cs:0.1152,35.6) {\textbf{0.098}};
\draw[white, line width=0.6pt, dash pattern=on 1.2pt off 1pt] (axis cs:0.1403,33) -- (axis cs:0.1403,43);
\node[f3lab, name=s2i, anchor=base, text=cSeekInk] at (axis cs:0.02105,10.6) {0.042};
\node[f3lab, name=s2o, anchor=base, text=white]    at (axis cs:0.0912,10.6) {\textbf{0.098}};
\node[f3lab, name=t0, fill=white, anchor=base west, text=cInk] at ([xshift=2.5pt]axis cs:0.1403,60.6) {\textbf{\$0.140}};
\node[f3lab, name=t1, fill=white, anchor=base west, text=cInk] at ([xshift=2.5pt]axis cs:0.1883,35.6) {\textbf{\$0.188}};
\node[f3lab, name=t2, fill=white, anchor=base west, text=cInk] at ([xshift=2.5pt]axis cs:0.1403,10.6) {\textbf{\$0.140}};
\node[f3lab, name=c1, fill=white, anchor=west, text=cInkB, align=left] at ([xshift=4pt]t1.east)
  {\textbf{\color{cSeekInk}2.1×} the output\\26.2k vs 12.4k tok};
\node[f3lab, name=c2, fill=white, anchor=base west, text=cInkB] at ([xshift=4pt]t2.base east)
  {same as ours, not below};
\node[f3lab, name=n0, fill=white, anchor=base west, text=cOursInk] at (axis cs:0,71.2) {\textbf{\method{} (ours)} {\color{cInkB}· 621 frames}};
\node[f3lab, name=n1, fill=white, anchor=base west, text=cSeekInk] at (axis cs:0,46.2) {\textbf{VideoSeek}, as it ran {\color{cInkB}· 123 frames}};
\node[f3lab, name=n2, fill=white, anchor=base west, text=cSeekInk] at (axis cs:0,21.2) {$\hookrightarrow$ repriced at cheapest image tier};
\draw[decorate, decoration={brace, amplitude=2.5pt, mirror}, cInkC, line width=0.5pt]
  ([xshift=-2.5pt]axis cs:0,68) -- ([xshift=-2.5pt]axis cs:0,33);
\node[f3lab, name=cIn, anchor=east, text=cInkB, align=right] at ([xshift=-7pt]axis cs:0,50.5)
  {same\\input};
\end{axis}
\path let \p1 = (pa.outer west), \p2 = (pa.north), \p3 = (pb.north), \p4 = (pb.west) in
  node[anchor=base west, font=\sffamily\footnotesize\bfseries, text=cInk, inner sep=0pt, name=ta]
    at (\x1, \y2+5pt) {(a) Accuracy vs.\ cost per question}
  node[anchor=base west, font=\sffamily\footnotesize\bfseries, text=cInk, inner sep=0pt, name=tb]
    at (\x4, \y3+5pt) {(b) Where the dollar goes};
\node[anchor=base east, font=\sffamily\scriptsize, text=cInkB, inner sep=0pt, name=key]
  at ($(pb.north east)+(0,5pt)$)
  {{\color{cOurs!28}\rule{3pt}{5.5pt}}{\color{cSeek!28}\rule{3pt}{5.5pt}}\,input\quad
   {\color{cOurs}\rule{3pt}{5.5pt}}{\color{cSeek}\rule{3pt}{5.5pt}}\,output};
\end{tikzpicture}%
\end{adjustbox}
  \caption{\textbf{Accuracy and cost on full LVBench} with Gemini~3.8 Flash. \textbf{(a)}~Measured accuracy versus cold-start API cost. \textbf{(b)}~Input/output cost breakdown on the same cost axis. The bottom bar reprices VideoSeek at the cheapest image tier while holding accuracy fixed; it is a pricing sensitivity analysis, not a measured configuration.}
  \label{fig:cost}
\end{figure}
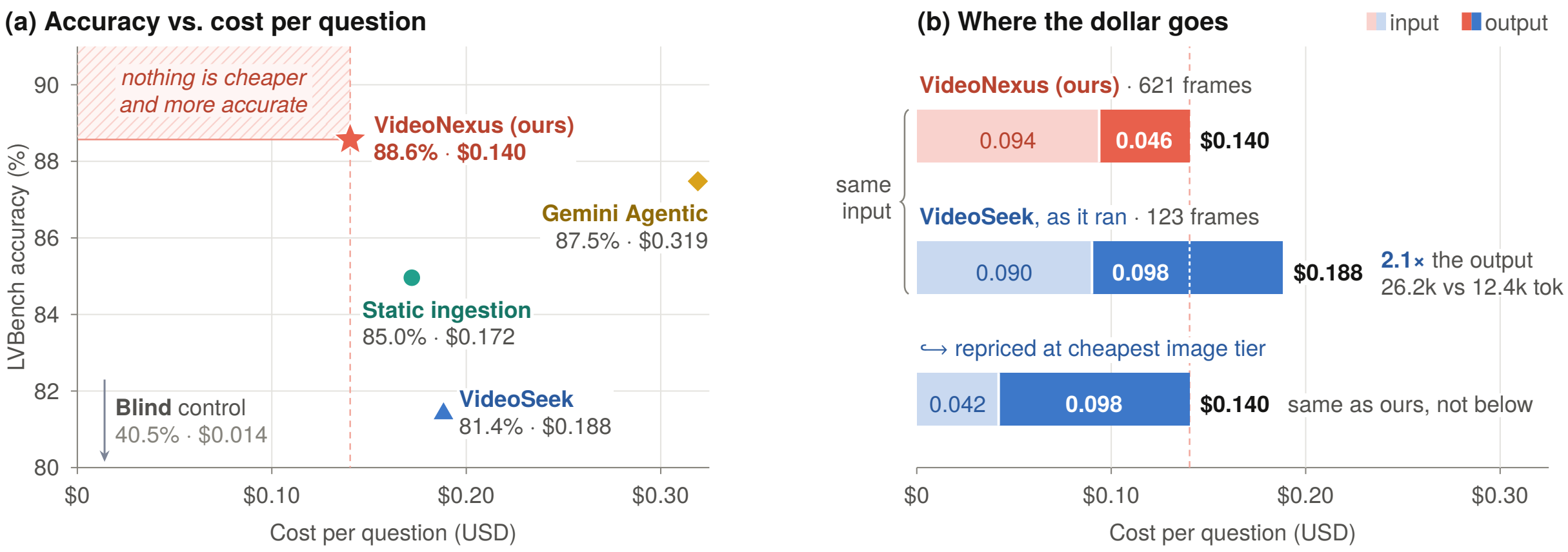

The lower block of \cref{tab:lvbench} compares \method with baselines that share the Gemini~3.8 Flash back-end.
On these hour-scale videos, Gemini Agentic improves over static ingestion by 2.5 points, but pays for it in the planner, with 72.4k reasoning tokens per question at 1.85$\times$ the cost of Static.
\method reaches \textbf{88.6\%} (1,372 of 1,549) at \$0.140 per question, and no method in our shared-backbone comparison is both cheaper and more accurate (\cref{fig:cost}a).
Against Static this is $+$3.6 points (90 questions won, 34 lost, exact McNemar $p{=}5{\times}10^{-7}$).
Against Gemini Agentic it is $+$1.1 points at 44\% of the cost; that difference, 57 won against 40 lost, is not significant ($p{=}0.10$). The observed accuracies are close, but this test does not establish equivalence.
The two read 621 and an estimated 464 frames per question, respectively. Their context organization and reasoning usage differ, but the closed baseline prevents isolating the source of this cost gap (\supp).

Against the VideoSeek re-run, which also delegates perception, \method is correct on 153 questions VideoSeek misses and wrong on 42 it answers (88.6\% against 81.4\%, $p{=}5{\times}10^{-16}$).
This is a system comparison: observation format, sampling, native audio access and media transport all differ.
The two use similar fresh input token counts per question, 118.2k for VideoSeek and 118.7k for \method (\cref{fig:cost}b), while processing 123 still-image frames and 621 native-video frames with audio, respectively.
They differ in generated output: VideoSeek emits 26.2k tokens per question against our 12.4k, and output is billed at five times the fresh-input price, accounting for most of the cost difference.
Repricing VideoSeek at the cheapest image tier gives \$0.140 per question; this is a pricing sensitivity analysis with accuracy held fixed, not another measured configuration (\supp).
\method also needs fewer model calls, 9.7 per question (6.1 planner turns, 3.6 perception calls) against 18.6.
It leads in five of six categories, by 2.7 points on reasoning and 3.6 on summarization; temporal grounding is the exception (92.3 against 90.8 for Gemini Agentic, one question).
Much of the remaining headroom is shared: 122 of \method's 177 errors are also errors of both Static and Gemini Agentic.

\subsection{Results on Video-GAIA}
\label{sec:exp:gaia}

\begin{table}[htbp]
  \centering
  \small
  \setlength{\tabcolsep}{3.2pt}
  \begin{tabular*}{\linewidth}{@{\extracolsep{\fill}}lccc|cccccc@{}}
    \toprule
    Method & All & L1 & L2 & Cul & Day & Eco & Geo & His & Tec \\
    \midrule
    Official ReAct & 51.7 & \textbf{79.0} & 43.5 & 48.3 & \textbf{67.2} & 33.3 & 41.3 & 54.5 & 48.7 \\
    \method & \textbf{60.2} & 77.4 & \textbf{55.0} & \textbf{55.2} & \textbf{67.2} & \textbf{50.0} & \textbf{50.8} & \textbf{68.2} & \textbf{64.1} \\
    \bottomrule
  \end{tabular*}
  \caption{\textbf{Video-GAIA results} (271 questions, LLM-judged accuracy). Both arms use Kimi~K3 as \planner and perception model and were run by us. Domains: Culture, Daily Life, Economics ($n{=}6$), Geography, History, Technology.}
  \label{tab:gaia}
\end{table}

With Kimi~K3 as \planner and perception model, \method answers 163 of 271 Video-GAIA questions against 140 for the official ReAct agent ($+$8.5 points; 51 won, 28 lost; \cref{tab:gaia}).
The gain is concentrated on L2 ($+$11.5), which requires locating a specific entity in the video and identifying it on the web; on L1 the two are within one question.
These questions were also used to develop the web and crop tools, so we read this as a transfer result on a different \planner rather than a calibrated margin (\supp).

\subsection{Delegated perception and visual memory}
\label{sec:exp:coupling}
\label{sec:exp:memory}

\begin{table}[htbp]
  \centering
  \small
  \setlength{\tabcolsep}{3.5pt}
  \begin{tabular*}{\linewidth}{@{\extracolsep{\fill}}lrrrr@{}}
    \toprule
    \multicolumn{5}{@{}l}{\emph{(a) Perception coupling: Gemini 3.8 Flash}} \\
    Local return & Acc. & Planner in & Planner out & \$/q \\
    \midrule
    Raw clip & 86.5 & 18.07 & 0.76 & 0.105 \\
    Raw clip, no Transcribe & 87.0 & 8.92 & 0.61 & 0.153 \\
    Text \finding & \textbf{89.0} & 12.33 & \textbf{0.57} & 0.110 \\
    \bottomrule
  \end{tabular*}
  \par\smallskip
  \begin{tabular*}{\linewidth}{@{\extracolsep{\fill}}lrrrr@{}}
    \toprule
    \multicolumn{5}{@{}l}{\emph{(b) Visual memory: text vs. selected frames}} \\
    Backbone & Text only & +local card & +global & Local $\Delta$ \\
    \midrule
    Gemini 3.0 Flash & 74.5 & \textbf{79.5} & 75.5 & +5.0 \\
    Gemini 3.8 Flash & 89.0 & \textbf{90.0} & -- & +1.0 \\
    \bottomrule
  \end{tabular*}
  \caption{\textbf{What should enter the reasoning context?} Both panels use dev-200. (a) Gemini~3.8 Flash coupling ablation; Transcribe and Overview are shared except in the marked no-Transcribe arm. Planner tokens are totals in millions. Per-call limits match; cumulative enforcement differs (\supp). (b) Card ablations within each backbone: local cards contain at most six frames; \texttt{+global} additionally returns Overview frames. Configurations differ across backbones, so gains do not isolate backbone effects.}
  \label{tab:coupling}
  \label{tab:memory}
\end{table}

The transcript-enabled pair holds model, runtime, skill, Overview and per-call limits fixed while changing the local observation interface (\cref{tab:coupling}).
In the coupled arms, \texttt{get\_frames} appends the native low-resolution audio--video interval to the \planner's context; in the decoupled arm, \texttt{inspect\_frames} returns a \finding from an independent call to the same model.
Decoupling reduces \planner input from 18.1M to 12.3M tokens ($-$32\%), while accuracy rises from 86.5\% (173/200) to 89.0\% (178/200) in this paired run. Total cost increases by 5\%, with perception processed in separate calls.
The 2.5-point accuracy increase comprises eight corrected answers and three regressions; this difference is not statistically significant (exact McNemar $p{=}0.23$).
Removing Transcribe from the coupled agent doubles Overview calls (92 $\rightarrow$ 187) and raises cost by 45\%: speech is a cheap localization cue.

Panel (a) disables evidence cards to isolate delegated local perception. Panel (b) asks whether selected visual evidence adds value to text reports: local cards raise Gemini~3.0 Flash from 74.5\% to 79.5\% and Gemini~3.8 Flash from 89.0\% (178/200) to 90.0\% (180/200). Both show gains, with a larger observed increase for Gemini~3.0; differing harness configurations prevent attributing that difference solely to backbone capability. Adding global Overview frames on top of local cards reduces Gemini~3.0 accuracy to 75.5\%. These results motivate retaining compact local visual memory alongside text, while delegating dense perception.

\subsection{Does an observation grow with what it perceived?}
\label{sec:exp:bounded}

\begin{figure}[htbp]
  \centering
  \includegraphics[width=0.54\linewidth]{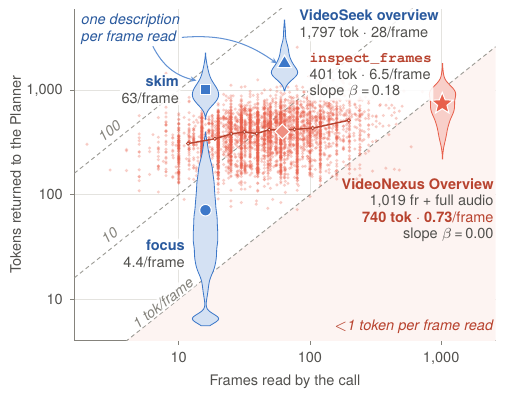}
  \caption{\textbf{Report size versus observed frames} on full LVBench (log--log axes). Coral dots are 4,376 Inspect calls with decile means; distributions summarize tools with nearly fixed frame counts. Dashed lines mark 1, 10, and 100 returned tokens per frame. Global scanning and local answering produce different report sizes, as discussed in the text.}
  \label{fig:bounded}
\end{figure}

We measure how strongly the size of a returned observation grows with the media inspected.
\Cref{fig:bounded} measures it on every perception call of the full run (per-tool table in \supp).
Overview consumes 141k media tokens on average, 1,019 frames plus the complete audio track, and returns 740 tokens, a 191$\times$ reduction, with a log--log slope of $0.00$ against perceived tokens ($r{=}0.01$), indicating little association within the observed range.
\texttt{inspect\_frames}, which is asked for detail, grows sub-linearly with slope $0.18$: between its lowest and highest decile of perceived media, frames rise 16.8$\times$ while returned text rises 1.67$\times$ (308 $\rightarrow$ 514 tokens).
Overview limits its candidate list, and Inspect answers a local task in a structured report. These measurements support weak output growth; they do not establish constant output size or calibrated confidence.
VideoSeek's scanning tools return approximately one description per frame: 63.8 descriptions per 64-frame \texttt{overview} and 14.4 per 16-frame \texttt{skim}, or 28 and 63 returned tokens per frame read. Its query-answering \texttt{focus} returns 4.4 tokens per frame and is also concise.
This baseline already uses concise local answers; the measured distinction is in the reports used for global navigation.
In the full run, perception reads 621 frames per question on average, while the \planner's context at its final turn holds 19.6k tokens on average, system prompt and skill included.

\subsection{Is sparse perception enough?}
\label{sec:exp:density}

\begin{table}[htbp]
  \centering
  \small
  \setlength{\tabcolsep}{3pt}
  \begin{tabular*}{\linewidth}{@{\extracolsep{\fill}}lr|ccc@{}}
    \toprule
    Global scan (frames per video) & Spacing & $\ge$1 frame & $\ge$2 frames & KIR $\ge$1 \\
    \midrule
    VideoXAgent budget (60.8) & 66.6\,s & 45.0 & 23.9 & 28.1 \\
    VideoSeek overview (64) & 63.2\,s & 45.9 & 24.5 & 28.7 \\
    VideoSeek budget (92.3) & 43.8\,s & 52.0 & 29.3 & 33.5 \\
    Uniform 384 & 10.5\,s & 72.8 & 54.6 & 49.4 \\
    \method Overview ($\le$1024) & 4.0\,s & 80.7 & 70.9 & 57.3 \\
    Static 1\,fps ($\le$2048) & 2.0\,s & 82.8 & 78.5 & 59.4 \\
    \bottomrule
  \end{tabular*}
  \caption{\textbf{How often does a uniform scan see the evidence?} Expected percentage of LVBench questions for which at least one (two) frames of a uniform scan with random phase fall inside the annotated evidence window, over 1,534 questions with a time reference. Median window 17\,s; 30\% of windows are at most 5\,s.}
  \label{tab:density}
\end{table}

LVBench annotates the time window that contains each question's evidence.
For a uniform scan with spacing $\Delta$, a window of length $d$ receives at least $k$ frames with probability $\mathrm{clip}(d/\Delta-k{+}1,0,1)$ (\cref{tab:density}).
At one frame per minute, a scan sees the evidence of only about half of the questions, and of 28\% of the key-information questions, which hinge on brief on-screen text or objects; VideoSeek reports that removing its overview costs 13.3 points~\cite{lin2026videoseek}, so the global scan carries much of the navigation.
A uniform scan at \method\textquotesingle s Overview budget is expected to place at least two frames inside 71\% of the annotated windows. This calculation uses uniform sampling, not either agent\textquotesingle s realized adaptive trajectory.
Overview is used on 41\% of questions; transcripts offer another localization route, so scan coverage is a discoverability proxy rather than an accuracy bound.

To measure density directly, we re-run the benchmark configuration on dev-200 with two numbers set to VideoSeek's budget: Overview capped at 64 frames and \texttt{inspect\_frames} at 16 frames per call.
Perception then reads 117 frames per question on average instead of 621, and accuracy falls from 90.0\% to 87.0\% (8 lost, 2 gained, $p{=}0.11$).
The reduced-budget run costs \$0.143 against \$0.140 per question, and uses more planner calls and output tokens: 8.0 planner calls instead of 6.1 and 15.3k output tokens instead of 12.4k.
The nominal accuracy difference is not significant; the cost figures also compare this subset run with the full-run average. They suggest a trade-off between observation size and subsequent interaction, rather than cost-free dense perception.
\method's lead over VideoSeek is also largest where the evidence window is short (\supp), the direction coverage predicts.

\subsection{Transfer across planner models}
\label{sec:exp:models}

\begin{table}[htbp]
  \centering
  \small
  \setlength{\tabcolsep}{3pt}
  \begin{tabular*}{\linewidth}{@{\extracolsep{\fill}}llccc@{}}
    \toprule
    Model & Benchmark & Baseline & \method & $\Delta$ \\
    \midrule
    Gemini 3.8 Flash & LVBench & 85.0 & \textbf{88.6} & +3.6 \\
    Gemini 3.0 Flash & LVBench & 74.0 & 80.4 & +6.4 \\
    Kimi K3 & LVBench$^\dagger$ & 73.0 & 79.5 & +6.5 \\
    Kimi K3 & Video-GAIA & 51.7 & 60.2 & +8.5 \\
    \bottomrule
  \end{tabular*}
  \caption{\textbf{Evaluation across models and configurations.} Baseline is static ingestion on LVBench and the official ReAct agent on Video-GAIA. $^\dagger$200-question subset.}
  \label{tab:models}
\end{table}

\Cref{tab:models} summarizes transfer experiments. Gemini~3.0 uses an earlier harness configuration, so these rows do not isolate the effect of changing the backbone (\supp).
\method improves Gemini~3.0 Flash by 6.4 points on full LVBench ($p<10^{-4}$), Kimi~K3 by 6.5 on a 200-question LVBench subset, and the official agent with Kimi~K3 by 8.5 on Video-GAIA.
The margin shrinks as the static baseline strengthens, from $+$6.4 at 74.0\% to $+$3.6 at 85.0\%.

\subsection{Failure modes}
\label{sec:exp:qual}

On the 153 questions VideoSeek misses and \method answers, VideoSeek takes 8.5 steps against its 6.2 average and scans globally first on 76\%. Continued interaction does not always resolve missing evidence; sampling and action-selection effects remain intertwined.
On the 42 that go the other way our trajectories run long, 9.7 turns against a typical 6.1, with Overview used on 29 of them and skipped on 13. Local inspection can therefore remain unsuccessful even after a global scan.
On Video-GAIA, 26 of 28 inspected cases find the right interval, while crops are readable enough to identify the target in 16 of 27 cases. This small qualitative sample points to spatial readability as a remaining difficulty.

\section{Conclusion}
\label{sec:conclusion}

Long-video agents must decide both what to observe and what to retain for reasoning.
\method supports this process through planner-directed perception in independent contexts and compact, temporally grounded reports for global navigation and local verification.
The paired ablation reduces planner input while increasing accuracy in that run, and per-call measurements show slowly growing reports across full LVBench.
With Gemini~3.8 Flash, it reaches 88.6\% on LVBench at 44\% of the built-in agentic mode's measured API cost. With Kimi~K3, it improves over the official Video-GAIA harness by 8.5 points in our evaluation.
These results support focused evidence exchange for video agents, subject to imperfect perception and the development and comparison limits discussed in \supp.
The visual-memory ablations motivate retaining selected evidence for verification alongside text reports. Future work should examine how observation density and retained evidence adapt to uncertainty across tasks and model capabilities.

\clearpage
{\small
\bibliographystyle{ieeenat_fullname}
\bibliography{main}}
\clearpage
\ifdefined\VNarxiv\else\setcounter{page}{1}\fi
\maketitlesupplementary
\appendix

The sections follow the order of the main paper: method details (\cref{sec:suppl:method}), baselines and cost accounting (\cref{sec:suppl:baseline,sec:exp:transport}), the per-tool observation sizes behind \cref{sec:exp:bounded} (\cref{sec:suppl:bounded}), coverage and density (\cref{sec:suppl:density}), visual memory (\cref{sec:suppl:memory}), the Video-GAIA protocol (\cref{sec:suppl:gaia}), qualitative analysis and complete trajectories (\cref{sec:suppl:qual,sec:suppl:traj}), further analyses, and limitations.

\section{Method details}
\label{sec:suppl:method}

\begin{figure*}[htbp]
  \centering
  \begin{adjustbox}{max width=\linewidth}
\input{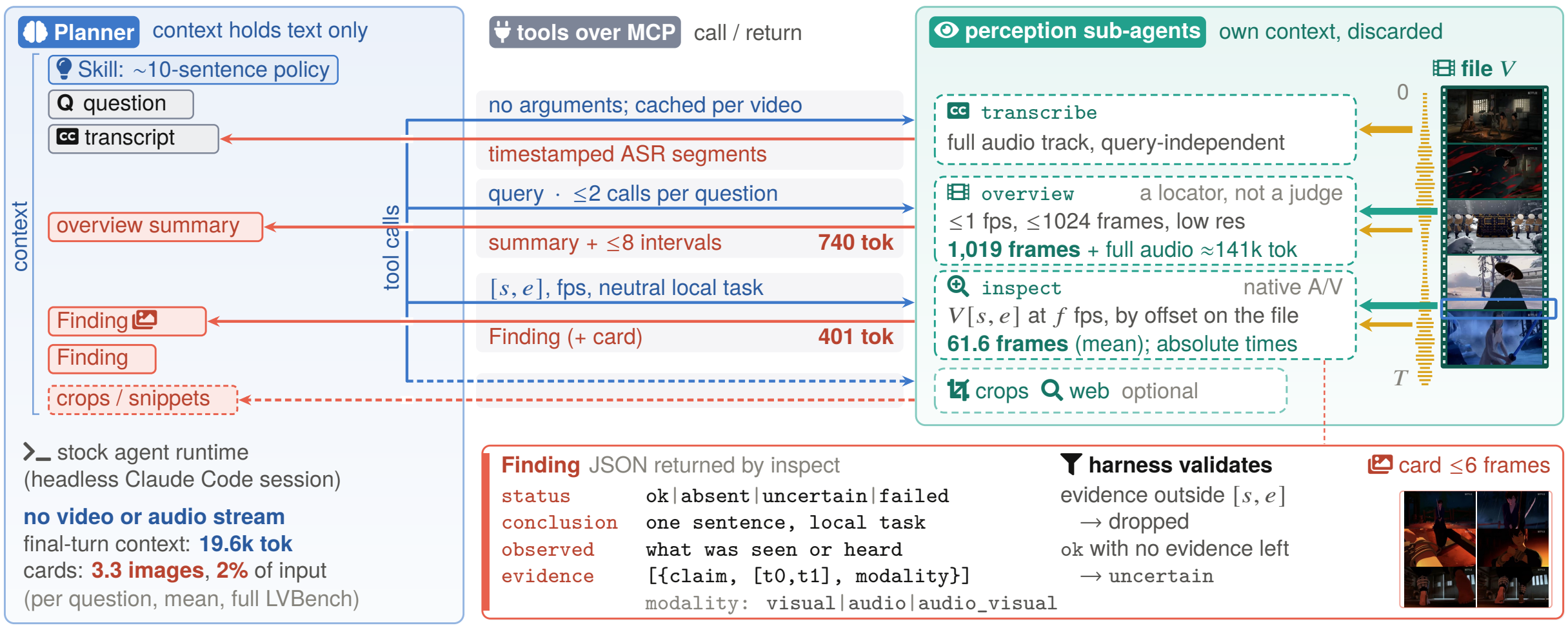}
\end{adjustbox}
  \caption{\textbf{Overview of \method.} The planner calls perception tools through MCP; each tool processes native media in an independent context and returns a compact report on the absolute source timeline. Thick arrows carry video and audio into perception; coral arrows carry reports back, optionally accompanied by evidence cards of at most six frames. Bottom: the \finding schema and validation. Dashed tools are optional; crops are used on both benchmarks, while web search and extraction are used only on Video-GAIA.}
  \label{fig:method}
\end{figure*}

\begin{table}[htbp]
  \centering
  \small
  \setlength{\tabcolsep}{3pt}
  \begin{tabular*}{\linewidth}{@{\extracolsep{\fill}}lll@{}}
    \toprule
    Tool & Planner input & Returned to \planner \\
    \midrule
    Transcribe & -- & timestamped ASR text \\
    Overview & query & summary, $\le$8 candidate intervals \\
    Inspect & $[s,e]$, fps, task & \finding (\cref{sec:method:finding}) \\
    \midrule
    \multicolumn{3}{@{}l}{\emph{Optional (crops also used on LVBench)}} \\
    Inspect Crops & time, regions, task & \finding on source-resolution crops \\
    Web Search / Extract & queries, URLs & snippets, page summaries \\
    \bottomrule
  \end{tabular*}
  \caption{\textbf{Tools available to the \planner.} All perception tools run an independent model call on native media. They return text, optionally with an evidence card of at most six frames anchored to the timestamps the \finding cites.}
  \label{tab:tools}
\end{table}

\Cref{fig:method} shows the harness schematically and \cref{tab:tools} lists the tools.
The skill imposes no fixed call order: the \planner may transcribe first, scan globally, inspect an interval named in the question directly, or answer immediately.
An evidence card adds limited visual context for checking a claim: at most six frames sampled at timestamps cited by the \finding.
Adding global-scan frames reduced accuracy by four points in the Gemini~3.0 development configuration (\cref{sec:suppl:memory}).
On full LVBench a \finding averages 401 tokens for 5.0k media tokens (62 frames plus audio), and between the lowest and highest decile of media perceived (12 and 198 frames) its text grows only 1.67$\times$.

\paragraph{Output and sampling limits.}
Overview truncates candidates to eight intervals; Inspect has no eight-item cap on its evidence array. Both use a maximum generation budget of 32,768 tokens, including thinking, and prompts encourage compact output. The frozen runners configure 256 frames per local call, ten calls and 2,048 cumulative frames. The direct-media implementation enforces all three; the delegated path bypasses the cumulative check. However, the largest recorded cumulative count is 1,024 across the available delegated development traces and 1,624 across the 1,549 saved full-run trajectories. Thus recorded delegated observations remain below the configured limit, despite the enforcement difference.

\paragraph{Earlier model configuration.}
The Gemini~3.0 result uses an earlier harness without ASR, combines the development subset and the remaining questions, and compares with a medium-resolution static baseline. It supports applicability beyond the main backbone but does not isolate a model swap in an otherwise identical harness.

\section{Baselines}
\label{sec:suppl:baseline}

\paragraph{Gemini Agentic.}
The mode was run with high thinking; 37 of 1,549 questions that failed repeatedly at high thinking were retried at medium.
It reports the media it navigated as \texttt{IMAGE} tokens under \texttt{toolUsePromptTokenCount}, never as video or audio, and for 92\% of the questions that count is an exact multiple of 66, the per-frame cost of low-resolution video, so the navigated frames enter the planning model's own context as images (\cref{sec:suppl:agentic}).
Dividing by 66 estimates 464 frames per question on average (median 150, maximum 12,106), which describes its observation volume without isolating the cause of its cost.

\paragraph{Reproducing VideoSeek.}
The baseline runs VideoSeek's released code with its own prompts, tools, step budget and sampling
configuration (\texttt{frame\_sampling\_factor} 4, \texttt{overview\_base} 16,
\texttt{skim\_base} 4, \texttt{focus\_base} 4, \texttt{max\_steps} 20,
\texttt{max\_tokens} 32768, temperature 1.0, seed 42): a 64-frame overview, 16-frame skims, and 16-second focus windows at 1\,fps in the tool description. We substitute the backbone and adapt the provider interface as detailed below.

Its agent asks the model for a plain-text thought and then makes a second, tool-enabled call to parse that text into actions, so the function calls in its history are synthesized locally rather than returned by the provider.
That design also asks for a plan in prose with no tools attached, and on this back-end an unconstrained turn answers such a prompt with a call to one of the provider's own built-in video tools instead, which reaches an OpenAI-style client as empty content: over half of the agent's steps then come back with an empty thought (57.6\% of its thoughts were lost when run that way), and its planner falls back to re-issuing the same overview until the step budget runs out.
Gemini~3 also attaches an opaque thought signature to each model turn, and replaying a function call without it is rejected.
We therefore drive the baseline through the native \texttt{streamGenerateContent} interface, declare function calling off on exactly the turns VideoSeek means to be tool-free, replay each model turn with the signature it was issued, and flatten the locally synthesized call envelope, whose action text each observation already restates verbatim.

We ran it at the high thinking level used by our own arms, above the \texttt{medium} its
configuration requests, to align thinking effort; this is a change from its default configuration. One asymmetry runs the other way and is
discussed under Limitations: it reads no transcript and no audio.

Two details of the released implementation matter for anyone repeating this. \texttt{focus}
advertises ``$\le \alpha{\cdot}4$\,s at 1\,FPS'' to the planner but caps on frame \emph{indices}
rather than seconds, so any window longer than about half a second yields the full
$\alpha{\cdot}4$ frames spread over it however long it is, which is why we take frames from the
configured budget rather than from the tool description. And its subtitle window filter combines
its two bounds with a disjunction where a conjunction is intended, making the filter a no-op for
any mid-video window.

\section{Cost accounting}
\label{sec:exp:transport}

\paragraph{Cold start.}
Static has cross-question cache hits because its video prefix repeats across questions on the same video; we reprice those hits as fresh input.
The two agentic rows have none: over all 1,549 questions, the first model call of a question is billed with zero cached tokens for both \method and the VideoSeek re-run, so every cache hit either one gets is a hit on its own question's history and is kept.
Pricing those within-question hits as fresh as well would read \$0.184 for \method and \$0.201 for VideoSeek, a 1.09$\times$ ratio instead of 1.34$\times$; the ordering survives either basis.

\paragraph{Sensitivity to media pricing.}
The provider bills the two transports differently: 66 tokens per frame for a native video stream at low resolution, against a per-image-part resolution tier for stills (264, 527, 1.1k and 2.2k prompt tokens per part, independent of the pixels in it) that VideoSeek leaves at its 1.1k-token default.
The measured cost therefore reflects both harness behavior and provider-specific media pricing.
Repricing every image part VideoSeek sent at each tier, and changing nothing else, gives \$0.252 per question at the most expensive tier, \$0.188 as it ran, \$0.155 one tier down and \$0.140 at the cheapest, against \$0.140 for \method.
At the cheapest tier the repriced costs are equal. Setting image charges to zero gives VideoSeek \$0.125 per question, 89\% of our cost; both calculations retain its measured accuracy.
The repricing holds accuracy fixed without validating that assumption: its \texttt{overview} packs eight frames into a single montage image, so the cheapest tier would leave those eight frames 264 tokens between them, and its \texttt{focus} sends frames at full resolution in order to resolve detail. An arm measuring the accuracy cost of that tier is the one reprice we have not run.
At the measured tier, the two harnesses use similar fresh input token counts, 118.2k fresh tokens per question against 118.7k, and differ 2.1$\times$ in output, which is priced at five times input; output is 52\% of VideoSeek's bill and 33\% of ours.
All dollar figures are list prices of one model and should not be carried to another provider.

\section{Observation size per tool}
\label{sec:suppl:bounded}

\begin{table}[htbp]
  \centering
  \small
  \setlength{\tabcolsep}{4pt}
  \begin{tabular*}{\linewidth}{@{\extracolsep{\fill}}lrrrrrr@{}}
    \toprule
    & & \multicolumn{2}{c}{Perceived} & \multicolumn{2}{c}{Returned} & \\
    \cmidrule(lr){3-4}\cmidrule(lr){5-6}
    Call & \#calls & frames & tokens & tokens & per frame & $\beta$ \\
    \midrule
    \multicolumn{7}{@{}l}{\emph{\method}} \\
    Overview        &   680 & 1{,}019 & 141{,}414 & 740 & 0.73 & 0.00 \\
    \texttt{inspect\_frames} & 4{,}376 &    61.6 &   5{,}046 & 401 & 6.5 & 0.18 \\
    \texttt{inspect\_crops}  &   448 &      -- &      --   & 189 & --          & -- \\
    \midrule
    \multicolumn{7}{@{}l}{\emph{VideoSeek~\cite{lin2026videoseek}, same model and provider}} \\
    \texttt{overview} & 1{,}257 & 64 &  9{,}998 & 1{,}797 & 28.1 & -- \\
    \texttt{skim}     & 2{,}777 & 16 & 17{,}775 & 1{,}011 & 63.2 & -- \\
    \texttt{focus}    & 3{,}915 & 16 & 17{,}730 &      71 & 4.4  & -- \\
    \bottomrule
  \end{tabular*}
  \caption{\textbf{Observation size against media perceived}, over all 5{,}504 perception calls of the full LVBench run. ``Perceived tokens'' are the \texttt{VIDEO}\,+\,\texttt{AUDIO} prompt tokens the call consumed; ``returned tokens'' are what the \planner receives, and ``per frame'' divides them by the frames the call read. $\beta$ is the log--log slope of returned against perceived tokens: 0 means the observation does not grow with the media at all, 1 means it grows proportionally. Crop calls read image parts rather than a video stream, so they report no media tokens.
VideoSeek sends still images and no audio, so its perceived tokens are what the provider bills for its image parts; its \texttt{overview} packs eight frames into each part, which is why its frames cost 156 tokens each against 1{,}108 for the two per-frame tools. Its rows are its 7{,}949 perception calls, over the 1{,}479 of the same 1{,}549 questions on which it made at least one and its call log aligns with its trajectory. Each of its tools reads a fixed number of frames per call, so its run has no variation to fit a slope to; the shape of the reply stands in for one: \texttt{overview} returns 63.8 frame descriptions per 64-frame call and \texttt{skim} 14.4 per 16-frame call, while \texttt{focus} answers the query and describes none.}
  \label{tab:bounded}
\end{table}

\Cref{tab:bounded} gives the per-tool numbers plotted in \cref{fig:bounded}.
Between its lowest and highest decile of perceived media, \texttt{inspect\_frames} frames rise 16.8$\times$ (11.8 $\rightarrow$ 197.7) while its returned text rises 1.67$\times$; Overview's report is uncorrelated with the media it read (Pearson $r{=}0.01$).
A reply made of one line per frame, as VideoSeek's \texttt{overview} and \texttt{skim} return, cannot stay fixed while its frame budget grows: every frame added to the scan is a line added to the \planner's context, re-read at every later step.
Its \texttt{focus} answers the query instead and returns 71 tokens for 16 frames.
The \planner of \method ends a question with a context of 19.6k tokens on average (median 17.0k), system prompt and skill included.

\section{Coverage and where density pays}
\label{sec:suppl:density}

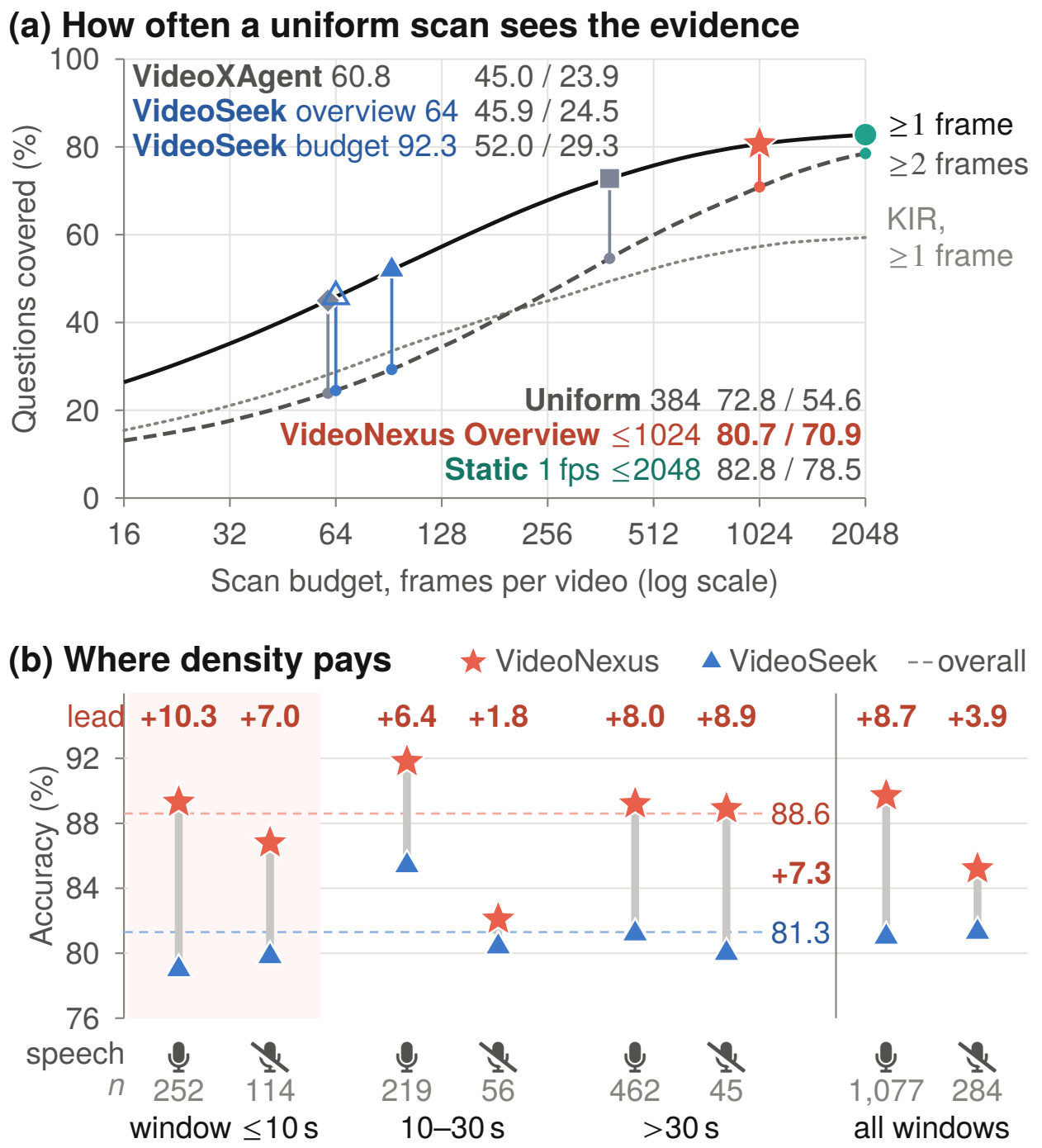
\begin{figure}[htbp]
  \centering
  \begin{adjustbox}{max width=\linewidth}
\pgfplotstableread[col sep=comma]{
N,ge1,ge2,kir
16.0,26.397,13.102,15.432
17.448,27.382,13.567,16.078
19.027,28.405,14.070,16.711
20.749,29.462,14.574,17.370
22.627,30.552,15.125,18.051
24.675,31.682,15.653,18.768
26.909,32.829,16.242,19.484
29.344,34.008,16.882,20.262
32.0,35.197,17.596,21.076
34.896,36.418,18.346,21.909
38.055,37.691,19.120,22.767
41.499,39.005,19.919,23.656
45.255,40.325,20.779,24.581
49.351,41.683,21.681,25.578
53.817,43.068,22.590,26.598
58.688,44.476,23.539,27.680
60.8,45.045,23.940,28.121
64.0,45.861,24.533,28.745
69.792,47.259,25.578,29.815
76.109,48.687,26.694,30.945
82.998,50.157,27.853,32.102
90.51,51.630,29.020,33.220
92.3,51.960,29.296,33.477
98.701,53.076,30.290,34.349
107.635,54.459,31.677,35.395
117.377,55.873,33.079,36.435
128.0,57.286,34.435,37.415
139.585,58.696,35.822,38.386
152.219,60.082,37.293,39.401
165.995,61.443,38.871,40.404
181.019,62.793,40.466,41.337
197.403,64.127,42.025,42.280
215.269,65.413,43.505,43.209
234.753,66.646,45.101,44.068
256.0,67.845,46.727,44.902
279.17,69.021,48.371,45.777
304.437,70.119,50.044,46.731
331.991,71.150,51.736,47.725
362.039,72.156,53.431,48.770
384,72.803,54.599,49.411
394.806,73.099,55.156,49.684
430.539,74.004,56.823,50.517
469.506,74.899,58.394,51.384
512.0,75.766,59.925,52.263
558.34,76.571,61.471,53.109
608.874,77.341,62.897,53.912
663.982,78.038,64.262,54.584
724.077,78.660,65.652,55.192
789.612,79.228,66.969,55.752
861.078,79.783,68.225,56.343
939.012,80.244,69.553,56.862
1024.0,80.665,70.867,57.310
1116.68,81.048,72.095,57.745
1217.748,81.378,73.304,58.146
1327.964,81.669,74.407,58.397
1448.155,81.924,75.395,58.620
1579.224,82.167,76.289,58.819
1722.156,82.399,77.166,58.986
1878.024,82.596,77.888,59.166
2048.0,82.776,78.527,59.363
}\fsixcov
\begin{tikzpicture}[
  f6lab/.style={font=\sffamily\scriptsize, inner sep=0.6pt},
  f6mk/.style={draw=white, line width=0.5pt, inner sep=0pt},
  f6star/.style={star, star points=5, star point ratio=2.3, minimum size=9pt, inner sep=0pt,
    fill=cOurs, draw=white, line width=0.45pt},
  f6tri/.style={regular polygon, regular polygon sides=3, minimum size=7pt, inner sep=0pt,
    fill=cSeek, draw=white, line width=0.45pt},
]
\pgfplotsset{f6base/.style={
  scale only axis, clip=false,
  axis lines*=left, axis line style={draw=cInkC, line width=0.4pt, -},
  tick align=outside, major tick length=2pt, tick style={draw=cInkC, line width=0.4pt},
  tick label style={font=\sffamily\scriptsize, text=cInkB},
  label style={font=\sffamily\scriptsize, text=cInkB},
  major grid style={draw=cGrid, line width=0.4pt},
}}
\begin{axis}[f6base, name=pa, width=169pt, height=100pt,
  xmode=log, log basis x=2, xmin=16, xmax=2048,
  xtick={16,32,64,128,256,512,1024,2048}, xticklabels={16,32,64,128,256,512,1024,2048},
  ymin=0, ymax=100, ytick={0,20,40,60,80,100}, yticklabels={0,20,40,60,80,100}, grid=major,
  ylabel={Questions covered (\%)}, ylabel shift=-3pt, ylabel style={inner sep=1pt},
  xlabel={Scan budget, frames per video (log scale)}, xlabel style={yshift=1.5pt}]
\addplot[cInkC, line width=0.7pt, dash pattern=on 0.6pt off 1.2pt, line cap=round] table[x=N, y=kir] {\fsixcov};
\addplot[cInkB, line width=0.9pt, dash pattern=on 3pt off 1.5pt] table[x=N, y=ge2] {\fsixcov};
\addplot[cInk, line width=0.9pt] table[x=N, y=ge1] {\fsixcov};
\node[f6lab, anchor=west, text=cInk]  at ([xshift=4pt,yshift=2.4pt]axis cs:2048,82.8) {${\ge}1$ frame};
\node[f6lab, anchor=west, text=cInkB] at ([xshift=4pt,yshift=-2.4pt]axis cs:2048,78.5) {${\ge}2$ frames};
\node[f6lab, anchor=west, text=cInkC, align=left] at ([xshift=4pt]axis cs:2048,59.4) {KIR,\\${\ge}1$ frame};
\def\fsixstem#1#2#3#4{\draw[#1, line width=0.7pt] (axis cs:#2,#4) -- (axis cs:#2,#3);
  \fill[#1] (axis cs:#2,#4) circle (1.3pt);}
\fsixstem{cSlate}{60.8}{45.0}{23.9}
\fsixstem{cSeek}{64}{45.9}{24.5}
\fsixstem{cSeek}{92.3}{52.0}{29.3}
\fsixstem{cSlate}{384}{72.8}{54.6}
\fsixstem{cOurs}{1024}{80.7}{70.9}
\fsixstem{cTeal}{2048}{82.8}{78.5}
\node[f6mk, diamond, fill=cSlate, minimum size=6pt] at (axis cs:60.8,45.0) {};
\node[f6tri, fill=none, draw=cSeek, line width=0.7pt, minimum size=6.5pt] at (axis cs:64,45.9) {};
\node[f6tri] at (axis cs:92.3,52.0) {};
\node[f6mk, rectangle, fill=cSlate, minimum size=4.8pt] at (axis cs:384,72.8) {};
\node[f6star] at (axis cs:1024,80.7) {};
\node[f6mk, circle, fill=cTeal, minimum size=5.4pt] at (axis cs:2048,82.8) {};
\node[f6lab, anchor=north west] at ([xshift=1pt]axis cs:16,100)
  {\setlength{\tabcolsep}{0pt}\renewcommand{\arraystretch}{0.95}%
   \begin{tabular}{l@{\enspace}l}
   {\color{cInkB}\textbf{VideoXAgent} 60.8} & {\color{cInkB}45.0 / 23.9}\\
   {\color{cSeekInk}\textbf{VideoSeek} overview 64} & {\color{cInkB}45.9 / 24.5}\\
   {\color{cSeekInk}\textbf{VideoSeek} budget 92.3} & {\color{cInkB}52.0 / 29.3}
   \end{tabular}};
\node[f6lab, anchor=south east] at (axis cs:2048,1)
  {\setlength{\tabcolsep}{0pt}\renewcommand{\arraystretch}{0.95}%
   \begin{tabular}{r@{\enspace}r}
   {\color{cInkB}\textbf{Uniform} 384} & {\color{cInkB}72.8 / 54.6}\\
   {\color{cOursInk}\textbf{\method{} Overview} ${\le}$1024} & {\color{cOursInk}\textbf{80.7 / 70.9}}\\
   {\color{cTealInk}\textbf{Static} 1\,fps ${\le}$2048} & {\color{cInkB}82.8 / 78.5}
   \end{tabular}};
\end{axis}
\begin{axis}[f6base, name=pb, at={(pa.south west)}, anchor=north west, yshift=-44.5pt,
  width=206pt, height=74pt, xmin=0.4, xmax=10.3, ymin=76, ymax=96,
  ytick={76,80,84,88,92}, yticklabels={76,80,84,88,92}, axis x line=none, grid=none,
  ylabel={Accuracy (\%)}, ylabel shift=-3pt, ylabel style={inner sep=1pt}]
\pgfplotsinvokeforeach{80,84,88,92}{\draw[cGrid, line width=0.4pt] (axis cs:0.4,#1) -- (axis cs:10.3,#1);}
\fill[cOurs!6] (axis cs:0.45,76) rectangle (axis cs:2.55,96);
\draw[cOurs!55, line width=0.5pt, dash pattern=on 2pt off 1.5pt] (axis cs:0.4,88.6) -- (axis cs:7.45,88.6);
\node[f6lab, anchor=west, text=cOursInk] at (axis cs:7.45,88.6) {88.6};
\node[f6lab, anchor=west, text=cOursInk, font=\sffamily\scriptsize\bfseries] at (axis cs:7.45,84.95) {+7.3};
\draw[cSeek!55, line width=0.5pt, dash pattern=on 2pt off 1.5pt] (axis cs:0.4,81.3) -- (axis cs:7.45,81.3);
\node[f6lab, anchor=west, text=cSeekInk] at (axis cs:7.45,81.3) {81.3};
\draw[cInkC, line width=0.4pt] (axis cs:8.2,76) -- (axis cs:8.2,96);
\def\fsixdb#1#2#3#4#5#6{%
  \draw[cInkC!45, line width=1.8pt] (axis cs:#1,#3) -- (axis cs:#1,#2);
  \node[f6tri] at (axis cs:#1,#3) {};
  \node[f6star] at (axis cs:#1,#2) {};
  \node[f6lab, anchor=center, text=cOursInk, font=\sffamily\scriptsize\bfseries] at (axis cs:#1,94.6) {+#4};
  \node[f6lab, anchor=north, text=cInkB] at ([yshift=-4.5pt]axis cs:#1,76) {#5};
  \node[f6lab, anchor=north, text=cInkC] at ([yshift=-13pt]axis cs:#1,76) {#6};}
\fsixdb{1}{89.3}{79.0}{10.3}{\faMicrophone}{252}
\fsixdb{2}{86.8}{79.8}{7.0}{\faMicrophoneSlash}{114}
\fsixdb{3.5}{91.8}{85.4}{6.4}{\faMicrophone}{219}
\fsixdb{4.5}{82.1}{80.4}{1.8}{\faMicrophoneSlash}{56}
\fsixdb{6}{89.2}{81.2}{8.0}{\faMicrophone}{462}
\fsixdb{7}{88.9}{80.0}{8.9}{\faMicrophoneSlash}{45}
\fsixdb{8.75}{89.7}{81.0}{8.7}{\faMicrophone}{1,077}
\fsixdb{9.75}{85.2}{81.3}{3.9}{\faMicrophoneSlash}{284}
\node[f6lab, anchor=north east, text=cInkB] at ([yshift=-4.5pt]axis cs:0.45,76) {speech};
\node[f6lab, anchor=east, text=cOursInk] at (axis cs:0.45,94.6) {lead};
\node[f6lab, anchor=north east, text=cInkC] at ([yshift=-13pt]axis cs:0.45,76) {\itshape n};
\node[f6lab, anchor=north, text=cInk] at ([yshift=-21.5pt]axis cs:1.5,76) {window ${\le}$10\,s};
\node[f6lab, anchor=north, text=cInk] at ([yshift=-21.5pt]axis cs:4,76) {10--30\,s};
\node[f6lab, anchor=north, text=cInk] at ([yshift=-21.5pt]axis cs:6.5,76) {${>}$30\,s};
\node[f6lab, anchor=north, text=cInk] at ([yshift=-21.5pt]axis cs:9.25,76) {all windows};
\end{axis}
\path let \p1 = (pa.outer west), \p2 = (pa.north), \p3 = (pb.north) in
  node[anchor=base west, font=\sffamily\footnotesize\bfseries, text=cInk, inner sep=0pt, name=ta]
    at (\x1, \y2+5pt) {(a) How often a uniform scan sees the evidence}
  node[anchor=base west, font=\sffamily\footnotesize\bfseries, text=cInk, inner sep=0pt, name=tb]
    at (\x1, \y3+5pt) {(b) Where density pays};
\node[anchor=base east, font=\sffamily\scriptsize, text=cInkB, inner sep=0pt, name=key]
  at ($(pb.north east)+(0,5pt)$)
  {\tikz[baseline=-0.6ex]\node[f6star, anchor=center, minimum size=7.5pt]{};\,\method\quad
   \tikz[baseline=-0.6ex]\node[f6tri, anchor=center, minimum size=6pt]{};\,VideoSeek\quad
   {\color{cInkC}\rule[0.5ex]{2.2pt}{0.5pt}\,\rule[0.5ex]{2.2pt}{0.5pt}}\,overall};
\end{tikzpicture}%
\end{adjustbox}
  \caption{\textbf{A sparse scan misses the evidence of about half the questions, and \method leads in every slice.}
  \textbf{(a)}~Expected share of the 1,534 LVBench questions with an annotated evidence window for which a uniform scan with random phase puts at least one (solid) or two (dashed) frames inside the window, against its budget in frames per video (capped at 1\,fps); dotted: at least one frame, key-information-retrieval (KIR) questions only. Stems mark the budgets of \cref{tab:density} and span their ${\ge}2$ and ${\ge}1$ values: at VideoSeek's 92.3 frames the scan sees the window for 52.0\% of questions, at \method's Overview for 80.7\%. The curves stay below 83.7\% because 250 windows are annotated as a single timestamp.
  \textbf{(b)}~Accuracy of \method (star) and VideoSeek (triangle) on the same questions, by the length of the evidence window and whether our transcript places speech inside it; \method's lead above each slice, $n$ below. Dashed lines: 88.6\% against 81.3\% (+7.3) over the 1,534 time-annotated questions; the main table uses all 1,549. The six slices hold the 1,148 questions with a window of positive length and a transcript; the right-hand columns pool all windows.}
  \label{fig:density}
\end{figure}

We compare the coverage calculation with descriptive performance slices from the two harnesses on the same questions (\cref{fig:density}).
We slice the 1{,}534 questions for which LVBench annotates an evidence window by the length of that window and by whether our cached transcript places speech inside it.
The six slices hold the 1{,}148 questions whose window has positive length and whose video has a cached transcript; another 213 questions with a transcript have a window annotated as a single timestamp and enter only the pooled speech/no-speech comparison.
\method leads in every one of the six slices, by between 1.8 and 10.3 points against an overall 7.3.
The single-timestamp questions are not uniformly favourable: on the 69 of them without speech VideoSeek is ahead (85.5\% against 82.6\%), while on the 144 with speech \method is (88.9\% against 77.1\%).

Two things separate the slices, and only one of them is density.
The lead is largest where the evidence window is short: 10.3 points on windows of at most 10\,s with speech and 7.0 on the 114 short windows without it, against 1.8 on 10--30\,s windows without speech, the direction coverage predicts, since a short window is what a sparse uniform scan steps over.
The other split is the lead of 8.7 points where speech falls inside the window (89.7 against 81.0 on 1{,}077 questions) against 3.9 where it does not (85.2 against 81.3 on 284).
Audio access differs between the systems, and the slices also differ in question content. This descriptive comparison cannot separate the contributions of density and modality.
These slices are descriptive and do not isolate a causal contribution of density.

\paragraph{The frugal-budget ablation.}
The density ablation of \cref{sec:exp:density} is paired on the question: all 200 dev questions at VideoSeek's budget against the same questions inside the full benchmark run, with 172 right in both.
The 90.0\% density control is the four-tool benchmark configuration with evidence cards, evaluated on these 200 questions within the full run. The coupling comparison in \cref{tab:coupling}a uses the earlier three-tool configuration without cards. Comparing those two configurations alone would not isolate the effect of cards. The visual-memory comparison is reported separately in \cref{tab:memory}b; its Gemini~3.8 row uses the aggregate counts described below.
Scan coverage bounds what a \emph{single} global scan can discover, not what the agent can answer, since speech and the transcript offer a second route to the same evidence; this is why we present coverage as a discoverability proxy and the ablation as the accuracy measurement.

\section{Visual memory in the planner's context}
\label{sec:suppl:memory}

\Cref{tab:memory}b reports the card ablations in the main paper. The Gemini~3.0 comparison holds the earlier harness and dev-200 questions fixed, changing whether local evidence cards and global Overview frames enter the planner. Local cards improve accuracy by 5 points; adding global frames reduces it by 4 points. The Gemini~3.8 comparison records 178 correct answers without cards and 180 with cards on dev-200, a 1-point gain. Only aggregate correct-answer counts are available for the Gemini~3.8 card comparison, so no discordant-pair counts or paired significance test are reported for it. These within-backbone results support compact local visual memory, but the earlier Gemini~3.0 harness differs from the Gemini~3.8 configuration and does not isolate a backbone effect.
In the coupling ablation, the nominal gain from decoupling is concentrated in entity recognition (58 $\rightarrow$ 62 of 72), but this category-level difference is descriptive; it rests on the same 200 questions.

\section{Video-GAIA protocol}
\label{sec:suppl:gaia}

Video-GAIA has six domains: Culture, Daily Life, Economics, Geography, History and Technology.
We score it with a binary semantic judge against the reference answers, with one deviation from the benchmark protocol: the official protocol polls a primary and a fallback judge and averages three runs, while we judge once with a single Kimi~K3 judge.
Both arms were run by us on the same 271 questions and scored by the same judge with the same
prompt and reference answers, so the paired comparison is internally consistent. The absolute
numbers are one draw of a stochastic judge -- 80 questions were judged at temperature 0 and 191 at
temperature 1 -- and should not be read against the public leaderboard.

Four settings are not aligned between the arms: the baseline is allowed 40 steps
against our 24 turns; our \planner runs with its thinking stripped from the replayed context while
the official agent keeps it; and the measured run predates a fix to the bounding-box coordinate
convention shared by the \planner and the crop tool, so its crop path was reading boxes in the
wrong frame. The fourth concerns observation access: the official agent perceives 20 uniformly sampled frames, while
ours reads a duration-preserving proxy video for the global pass and native intervals locally.
Perception usage records are missing for 291 of 1,342 calls on our side, so we report no cost
comparison on this benchmark.
There is also development overlap: these 271 questions were also the set on which we developed the web-search tools and the crop protocol, so for \method they are not held out, and we read \cref{tab:gaia} as a transfer result and a direction rather than a calibrated margin.

\section{Qualitative analysis}
\label{sec:suppl:qual}
Of the 37 v0 errors on dev-200, 15 stem from localization or fine-grained identity, 8 from counting or ordering across segments, and 3 from mapping correct evidence to the wrong option.
On the 153 questions VideoSeek misses and \method answers, VideoSeek took 8.5 steps against its own 6.2 average, called \texttt{focus} 630 times, and had scanned globally first on 76\% of them: these traces show continued evidence seeking but do not identify sampling as the sole cause of failure.
On the 42 that go the other way, our trajectories make 214 \texttt{inspect\_frames} calls in total and use Overview on 29 questions. Detailed local inspection can remain unsuccessful even after global navigation.
\Cref{fig:trajectory} in the main paper follows one question (\texttt{lvbench\_10}) through both kinds of call; \cref{sec:suppl:traj} gives five further complete trajectories, including a failure, and the behaviour of the loop over the full run.


\newcommand{\trnum}[1]{\tikz[baseline=(n.base)]\node[circle,draw=cPlan,fill=white,line width=.5pt,inner sep=0pt,minimum size=9pt,font=\sffamily\scriptsize,text=cPlanInk](n){#1};}
\newcommand{\trtool}[1]{\textcolor{cPlanInk}{\texttt{\textbf{#1}}}}
\newcommand{\trread}[1]{\enspace\textcolor{cTealInk}{\sffamily\scriptsize$\triangleright$\,#1}}
\newcommand{\trthink}[1]{\textcolor{cInkB}{\textit{\planner thought:} ``#1''}}
\newcommand{\trfind}[2]{\textcolor{cOursInk}{\sffamily\bfseries\finding\,\texttt{\mdseries#1}}\enspace#2}
\newcommand{\trnote}[1]{\textcolor{cInkC}{\sffamily[#1]}}
\newcommand{\trok}{\textcolor{cTealInk}{\cmark}}
\newcommand{\trno}{\textcolor{cOursInk}{\xmark}}
\newenvironment{trsteps}{\begin{list}{}{\setlength{\leftmargin}{14pt}\setlength{\labelwidth}{10pt}%
  \setlength{\labelsep}{4pt}\setlength{\itemsep}{2.5pt}\setlength{\topsep}{3pt}\setlength{\parsep}{0pt}}}{\end{list}}
\newcommand{\trstep}[1]{\item[\trnum{#1}]}
\newcommand{\trhead}[5]{\noindent\colorbox{cSlate!9}{\parbox{\dimexpr\textwidth-2\fboxsep\relax}{\footnotesize
  \textbf{Q.}~#3\hfill\textcolor{cInkB}{\texttt{#1}\,\textperiodcentered\,#2}\\[1pt]#4\\[1pt]#5}}}

\section{Complete trajectories}
\label{sec:suppl:traj}

\begin{figure*}[p]
\trhead{lvbench\_1248}{event understanding, 36-min vlog, annotated 17:30--19:20}%
  {How many pieces of sandwiche does the vlogger eat?}%
  {(A) Two\quad (B) One\quad (C) Three\quad (D) All}%
  {Ground truth \textbf{A}.\quad \method~A\,\trok\quad Gemini Agentic~B\,\trno\quad Static~A\,\trok\quad VideoSeek~B\,\trno\hfill\textcolor{cInkB}{4 \planner turns, 8.5k \planner input tokens}}
\par\vspace{4pt}
\begin{minipage}[t]{0.625\textwidth}\footnotesize\raggedright
\begin{trsteps}
\trstep{1}\trtool{get\_transcript}() \trnote{no transcript line between 18:20 and 19:30}
\trstep{2}\trtool{overview}(query ``eating sandwich sandwiches vlogger eats sandwich'')\trread{1,024 frames + audio track: 67,584 video + 53,361 audio tokens}\\
  \trfind{}{``The vlogger prepares a multi-layer toasted club sandwich [\ldots] around 14:37--18:17. She serves it with iced coffee and eats the sandwich at the dining table from approximately 18:44 to 19:07 before packing up leftovers.''}
  Candidates: 18:43--19:08 ``[\ldots] picks up a club sandwich quarter, removes the skewer picks, and eats it [\ldots]''; 18:00--18:43 ``cuts the sandwich into quarters, plates it [\ldots]''; 19:08--19:25 ``transfers the remaining sandwich pieces and fries into a collapsible food storage box.''
\trstep{3}\trtool{inspect\_frames}(1120--1170\,s, 1\,fps, low; task ``Count how many pieces of sandwich the vlogger eats, and observe how many pieces remain on the plate and are packed into the storage container.'')\trread{50 frames + audio: 3,300 video + 1,250 audio tokens}\\
  \trfind{ok}{``The vlogger eats 2 pieces of sandwich, while the remaining 2 pieces on the plate are packed into the pink storage container.''}
  E1 [1126--1128\,s] ``The plate displays 4 sandwich pieces, and the vlogger picks up the first piece.''
  E2 [1129--1131\,s] ``The vlogger eats the first piece of sandwich.''
  E3 [1143--1146\,s] ``The vlogger picks up and eats the second sandwich piece from the brown plate.''
  E4 [1149--1153\,s] ``The remaining 2 sandwich pieces are transferred from the plate into the pink storage container.''
\trstep{4}Answer \textbf{A} \trok
\end{trsteps}
\end{minipage}\hfill
\begin{minipage}[t]{0.355\textwidth}\vspace{3pt}
\includegraphics[width=\linewidth]{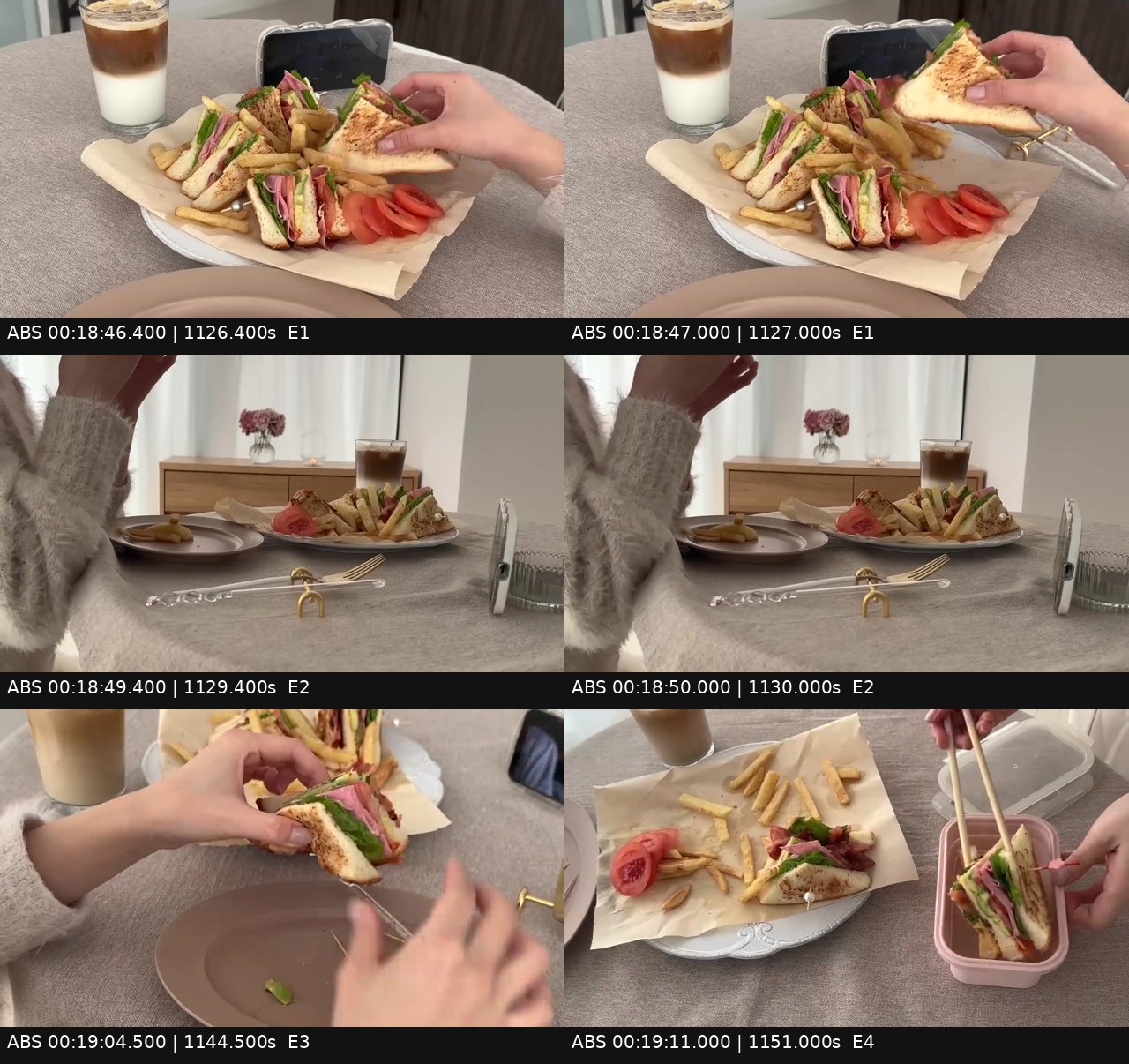}\\[1pt]
{\scriptsize\sffamily\color{cInkB} Evidence card returned with the \finding of step~3, as the \planner received it: 6 of the 50 frames, tagged with claims E1--E4.}
\end{minipage}
\caption{\textbf{Complete trajectory: counting.} One Overview locates the meal; one Inspect over the eating and the packing counts both. Annotations as in \cref{sec:suppl:traj}; teal $\triangleright$ is what the perception sub-agent read for the call.}
\label{fig:traj:1248}
\end{figure*}

\begin{figure*}[p]
\trhead{lvbench\_734}{event understanding, 92-min film without dialogue, annotated 1:08:07--1:08:26}%
  {One time a smaller insect saves a larger insect of the same class in the rainforest, what happens?}%
  {(A) The smaller ladybug saves the larger ladybug from a spider's web\quad (B) The smaller mantis saves the larger mantis from a human's hand\\(C) The smaller ladybug saves the larger ladybug from a mantis's hand\quad (D) The smaller mantis saves the larger mantis from a spider's web}%
  {Ground truth \textbf{C}.\quad \method~C\,\trok\quad Gemini Agentic~A\,\trno\quad Static~A\,\trno\quad VideoSeek~A\,\trno\hfill\textcolor{cInkB}{12 \planner turns, 113.4k \planner input tokens}}
\par\vspace{4pt}
\begin{minipage}[t]{0.625\textwidth}\footnotesize\raggedright
\begin{trsteps}
\trstep{1}\trtool{get\_transcript}() \trnote{23 lines, none of them dialogue, \eg ``Sous rique?'', ``Musique?''}
\trstep{2}\trtool{overview}(query ``smaller insect saves larger insect mantis ladybug spider web'')\trread{1,024 frames + audio track: 61,440 video + 138,351 audio tokens}\\
  \trfind{}{``[\ldots] In the Guadeloupe jungle (\textasciitilde43:05--49:25), a praying mantis chases a small ladybug, causing it to fall into an underground cave featuring a colossal spider web [\ldots]. Later (\textasciitilde67:35--70:50), praying mantises hunt and corner the ladybugs on a cliff [\ldots]''} Seven candidates, among them 43:55--45:10 and 67:35--69:15.
\trstep{3}\trtool{inspect\_frames}(2640--2720\,s, 1\,fps, low)\trread{80 frames + audio: 4,800 video + 2,000 audio tokens}\\
  \trfind{ok}{``A baby ladybug, a fly, and an adult ladybug are trapped in the spider web. [\ldots]''}
\trstep{4}\trtool{inspect\_frames}(2720--2860\,s, 1\,fps, low; task ``Observe what happens to the trapped ladybugs, the spider, and who saves whom.'')\trread{140 frames + audio: 8,400 video + 3,500 audio tokens}\\
  \trfind{ok}{``A trapped ladybug stuck on a web is saved by a flying ladybug, who drops a heavy stone on the spider to knock it down and then frees the trapped companion so they can escape together.''}
\trstep{5}\trtool{inspect\_frames}(2780--2810\,s, 1\,fps, high; task ``Identify which ladybug is trapped (larger adult ladybug or smaller baby ladybug) and which ladybug saves it (smaller or larger). Check sizes and appearances.'')\trread{30 frames + audio: 7,500 video + 750 audio tokens}\\
  \trfind{ok}{``The trapped insect is the smaller baby ladybug, while the rescuer is the larger adult ladybug.''} \trnote{the reverse of what (A) needs}
\trstep{6}\trthink{My current focus is on dissecting the provided conclusion against the question's premise. I'm scrutinizing the apparent contradiction between the smaller insect being the rescuer and the larger insect being the trapped one [\ldots]}\\
  \trtool{inspect\_frames}(2635--2675\,s, 1\,fps, high)\trread{40 frames: 10,000 video + 1,000 audio tokens}\enspace
  \trfind{ok}{``[\ldots] Both the adult ladybug and a baby ladybug are trapped on the spiderweb.''}
\trstep{7}\trtool{inspect\_frames}(2710--2765\,s, 1\,fps, low)\trread{55 frames: 3,300 video + 1,375 audio tokens}\enspace
  \trfind{ok}{``[\ldots] the trapped ladybug remains stuck on the web through 2765.''}
\trstep{8}\trtool{inspect\_frames}(2760--2805\,s, 1\,fps, high; task ``Look at the two ladybugs: which one is trapped on the web when the spider dances, and which one drops the rock from above? [\ldots]'')\trread{45 frames: 11,250 video + 1,125 audio tokens}\\
  \trfind{ok}{``The smaller ladybug is trapped on the web while the spider dances, whereas the larger ladybug is perched on the rock above and releases/drops it to knock the spider away. [\ldots]''}
\trstep{9}\trthink{[\ldots] I now see the question specifies a smaller insect saving a larger one of the same class, prompting a re-examination of the options [\ldots]}\\
  \trtool{overview}(query ``mantis human hand spider web saves smaller mantis larger mantis'')\trread{1,024 frames + audio track: 61,440 video + 138,351 audio tokens}\\
  \trfind{}{``[\ldots] no single sequence simultaneously combines a human hand, a spider web, and a mantis saving or interacting with another mantis [\ldots]''} Six candidates, among them 67:45--70:45.
\trstep{10}\trtool{inspect\_frames}(4060--4120\,s, 1\,fps, high; task ``What happens between the praying mantis and the ladybugs? Does the mantis catch a ladybug in its hand/claws? Does another ladybug save it? Which ladybug is smaller and which is larger?'')\trread{60 frames + audio: 15,000 video + 1,500 audio tokens}\\
  \trfind{ok}{``The praying mantis catches a ladybug with its front raptorial claws, but another ladybug flies in at high speed and knocks it free, saving it. The ladybug that was caught is the larger one, while the ladybug that rescues it is smaller.''}
\trstep{11}Answer \textbf{C} \trok
\end{trsteps}
\end{minipage}\hfill
\begin{minipage}[t]{0.355\textwidth}\vspace{3pt}
\includegraphics[width=\linewidth]{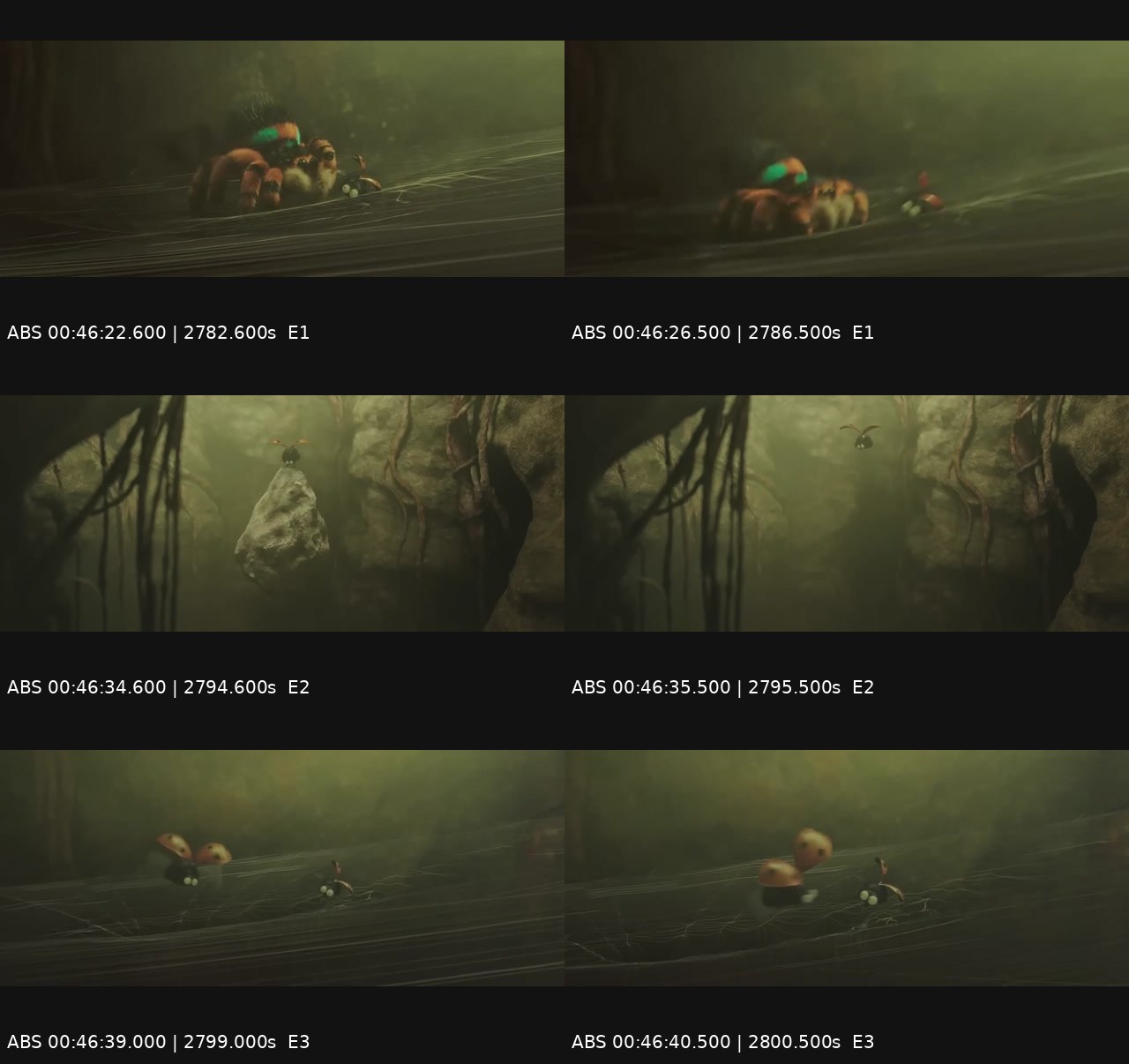}\\[1pt]
{\scriptsize\sffamily\color{cInkB} Evidence card of step~5 (46:22--46:40): the small ladybug with one red wing cover is caught in the web; the larger one drops the stone and lands beside it.}\\[5pt]
\includegraphics[width=\linewidth]{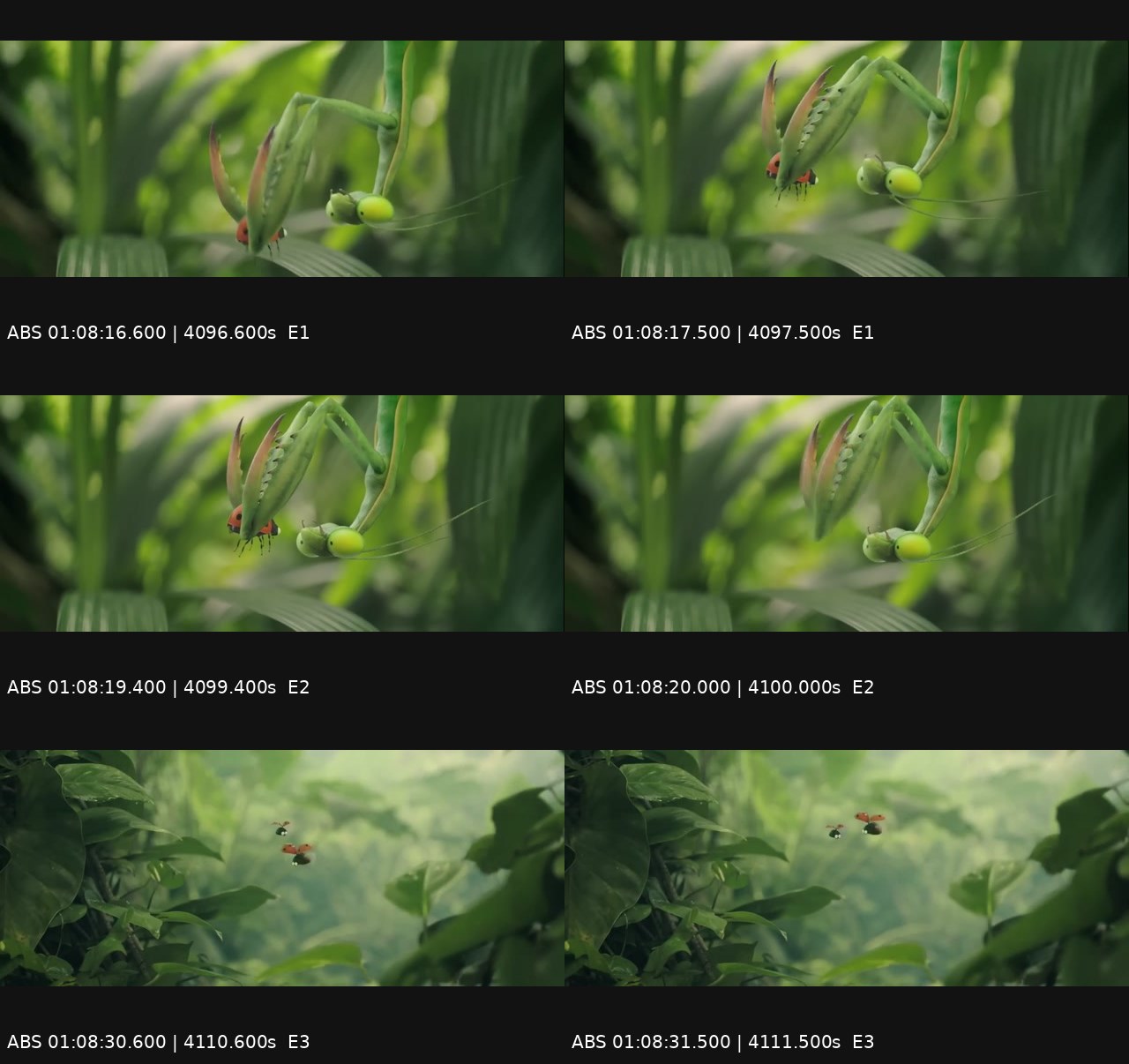}\\[1pt]
{\scriptsize\sffamily\color{cInkB} Evidence card of step~10 (1:08:16--1:08:31): the mantis holds a ladybug; it is knocked free, and a small and a large ladybug fly off together.}
\end{minipage}
\caption{\textbf{Complete trajectory: a \finding that contradicts the expected option.} The web rescue looks like option (A), the answer of all three baselines, but the high-resolution \finding{}s report that there the larger ladybug saves the smaller one. The \planner re-runs Overview with a new query and finds the scene that fits.}
\label{fig:traj:734}
\end{figure*}

\begin{figure*}[p]
\trhead{lvbench\_598}{key information retrieval, 60-min compilation, annotated 03:49}%
  {What is the name of the song sung by the third contestant?}%
  {(A) Creep\quad (B) Savage Love (Laxed -- Siren Beat)\quad (C) Turning Tables\quad (D) Break My Heart Again}%
  {Ground truth \textbf{A}.\quad \method~A\,\trok\quad Gemini Agentic~A\,\trok\quad Static~A\,\trok\quad VideoSeek~A\,\trok\hfill\textcolor{cInkB}{6 \planner turns, 59.6k \planner input tokens}}
\par\vspace{4pt}
\begin{minipage}[t]{0.66\textwidth}\footnotesize\raggedright
\begin{trsteps}
\trstep{1}\trtool{get\_transcript}() \trnote{lyrics of three songs, e.g.:} [41.6--49.6] ``To make you stay, but you're savage.'' [186.9--193.1] ``Go ahead and break my heart again, believe me.'' [214.2--219.2] ``And you were here before.''
\trstep{2}\trthink{I've successfully identified three distinct musical pieces within the transcript. My current focus is on confirming the lyrical associations and ensuring accurate song attributions for each segment.}\\
  \trtool{inspect\_frames}(200--240\,s, 1\,fps, low; task ``Who is singing around 210-240 seconds, what song is it, and is this the 3rd contestant? Look for captions/text introducing the contestant and song title.'')\trread{40 frames + audio: 2,640 video + 1,000 audio tokens}\\
  \trfind{uncertain}{``Between 210 and 240 seconds, Max Novik is singing `Creep' by Radiohead, as identified by the on-screen text shown between 227 and 230 seconds. However, it is uncertain whether he is the third contestant, as neither the caption nor the audio indicates any contestant numbering.''}
  E3 [200--214\,s] ``The preceding performance from The Voice Norway concludes and transitions into a bumper without showing any overall contestant counter.''
\trstep{3}\trthink{I'm currently cross-referencing estimated durations for contestants one and two, aiming for precise start and end points. [\ldots]}\\
  \trtool{inspect\_frames}(0--30\,s, 1\,fps, low; task ``Who is the first contestant and what song are they singing? Look for on-screen lower third title.'')\trread{30 frames + audio: 1,980 video + 750 audio tokens}\\
  \trfind{ok}{``The first contestant is Etienne Steven, and she is singing `Savage Love' (by Jawsh 685 \& Jason Derulo).''} \trnote{option (B)}
\trstep{4}\trtool{inspect\_frames}(95--125\,s, 1\,fps, low; task ``Who is the second contestant and what song are they singing? Look for on-screen lower third title.'')\trread{30 frames + audio: 1,980 video + 750 audio tokens}\\
  \trfind{ok}{``The contestant is Isak \O{}vrevold, and he is singing `Break My Heart Again' (originally by Finneas).''} \trnote{option (D)}
\trstep{5}Answer \textbf{A} \trok
\end{trsteps}
\end{minipage}\hfill
\begin{minipage}[t]{0.3\textwidth}\vspace{3pt}
\includegraphics[width=\linewidth]{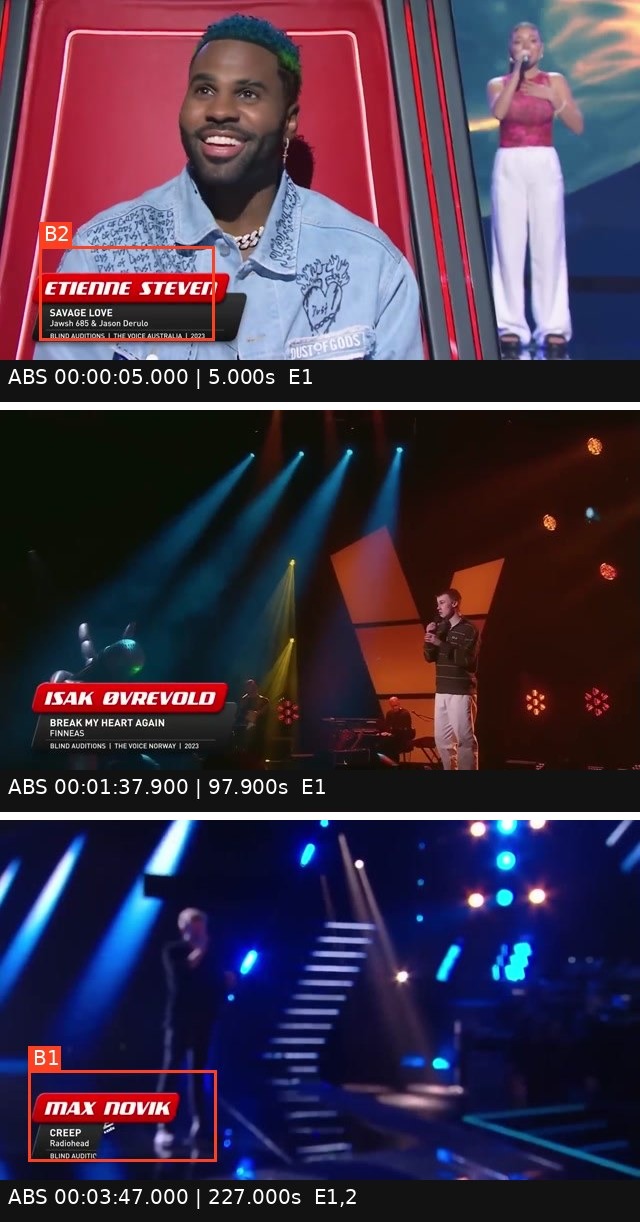}\\[1pt]
{\scriptsize\sffamily\color{cInkB} One frame from each evidence card, in video order: steps~3 (0:05), 4 (1:38) and 2 (3:47). Each caption names the singer, the song and the edition.}
\end{minipage}
\caption{\textbf{Complete trajectory: several occurrences, one asked for.} The first Inspect names the song but reports \texttt{uncertain} on whether its singer is the third contestant, since nothing on screen numbers them. Two cheap Inspects read the first two captions; their songs are the distractors (B) and (D).}
\label{fig:traj:598}
\end{figure*}

\begin{figure*}[p]
\trhead{lvbench\_1087}{temporal grounding, 84-min concert, annotated 05:45--05:46}%
  {How many TV hoods appeared after 05:45?}%
  {(A) 7\quad (B) 3\quad (C) 4\quad (D) 1}%
  {Ground truth \textbf{B}.\quad \method~B\,\trok\quad Gemini Agentic~C\,\trno\quad Static~D\,\trno\quad VideoSeek~B\,\trok\hfill\textcolor{cInkB}{10 \planner turns, 144.2k \planner input tokens}}
\par\vspace{4pt}
\begin{minipage}[t]{0.625\textwidth}\footnotesize\raggedright
\begin{trsteps}
\trstep{1}\trtool{get\_transcript}() \trnote{at the annotated time: [343.4--346.4] ``Change to the rhythm.''}
\trstep{2}\trthink{I've confirmed the dancers in ``Chained to the Rhythm'' do indeed wear TV sets as headgear, often termed ``TV hoods,'' which aligns perfectly with the transcript's timing and song cues. [\ldots]}\\
  \trtool{inspect\_frames}(340--400\,s, 1\,fps, low; task ``Look for dancers or people wearing TV hoods / TV headpieces / TVs on their heads. Count how many TV hoods appear after 05:45.'')\trread{60 frames + audio: 3,960 video + 1,500 audio tokens}\\
  \trfind{ok}{``After 05:45, exactly 4 TV hoods appear during the finale of the song from 05:45 to 05:46. [\ldots]''}
  B1 [346\,s] ``Four TV hoods are visible framing Katy Perry at 05:46.'' \trnote{the box holds three}
\trstep{3}\trtool{inspect\_frames}(344--350\,s, 2\,fps, high; task ``Count precisely how many TV hoods (TV headpieces / dancers wearing TV heads) appear at or after 05:45 (345.0s to 350.0s). List the count visible at each second.'')\trread{12 frames + audio: 3,168 video + 150 audio tokens}\\
  \trfind{ok}{``[\ldots] at 05:45 (345.0s), 1 TV hood is visible; at 05:46 (346.0s), 3 TV hoods are visible; and from 05:47 to 05:50 (347.0s to 350.0s), 0 TV hoods are visible.''}
\trstep{4}\trthink{[\ldots] My attention is drawn to a detail mentioned in the initial inspection: four TV hoods framing Katy Perry. [\ldots]}\\
  \trtool{inspect\_crops}(region B1; task ``Count exactly how many TV hoods (TV screens/monitors on dancers' heads) are framing Katy Perry or present in the scene at 346.0s. Look carefully at all dancers around Katy Perry.'')\trread{one source-resolution crop: 2,664 prompt tokens}\\
  \trfind{ok}{``There are exactly 3 TV hoods framing Katy Perry in the scene.''}
\trstep{5}\trtool{inspect\_frames}(315--345\,s, 0.5\,fps, low; task ``Observe the dancers wearing TV hoods during the performance of Chained to the Rhythm. How many dancers with TV hoods are there on stage throughout the song? Look at wide shots of the full stage.'')\trread{15 frames + audio: 990 video + 750 audio tokens}\\
  \trfind{ok}{``There are 4 dancers wearing TV hoods on stage.''} \trnote{evidence at 05:30 and 05:33}
\trstep{6}\trtool{inspect\_frames}(343--347\,s, 2\,fps, high; task ``Track the 4 TV-hooded dancers from 343 to 347. Did one leave, or where did each of them go? How many TV hoods are visible on screen after 05:45 (345.0s)?'')\trread{8 frames + audio: 2,112 video + 100 audio tokens}\\
  \trfind{ok}{``Only 3 TV-hooded dancers are present, not 4. [\ldots] After 05:45, exactly 3 TV hoods are visible on screen.''}
\trstep{7}\trthink{[\ldots] the subsequent wide shots between 345.5 and 346.5 seconds, showing exactly three TV hoods, are consistent. No other songs in the setlist utilize this prop.}\\
  \trtool{inspect\_frames}(345--347\,s, 4\,fps, high; task ``Count the exact number of TV hoods visible after 05:45.0. Are there 3 or 4 distinct TV hoods visible at any point? Check every frame carefully.'')\trread{8 frames + audio: 2,112 video + 50 audio tokens}\\
  \trfind{ok}{``There are exactly 3 distinct TV hoods visible after 05:45.0, not 4.''}
\trstep{8}Answer \textbf{B} \trok
\end{trsteps}
\end{minipage}\hfill
\begin{minipage}[t]{0.355\textwidth}\vspace{3pt}
\includegraphics[width=\linewidth]{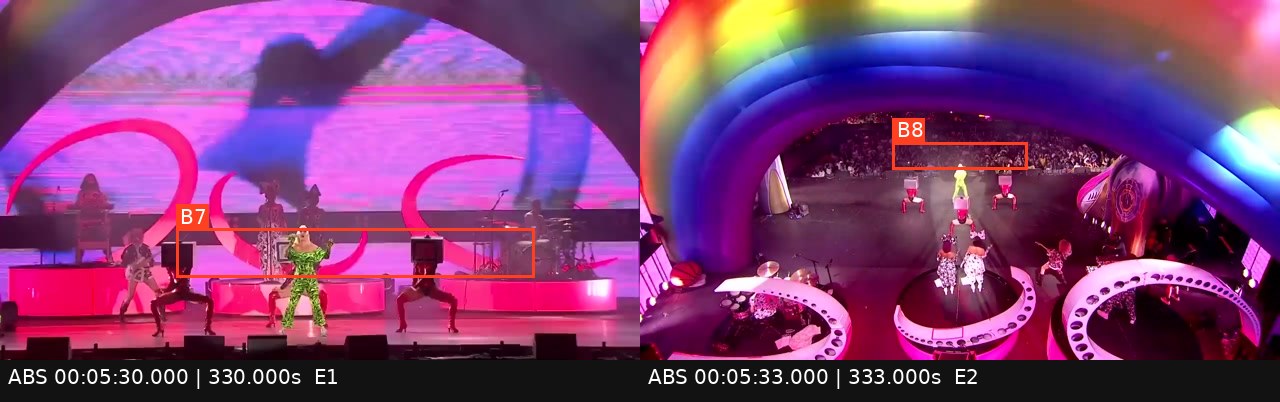}\\[1pt]
{\scriptsize\sffamily\color{cInkB} Evidence card of step~5 (05:30, 05:33), earlier in the song, on which it counts four hooded dancers.}\\[5pt]
\includegraphics[width=\linewidth]{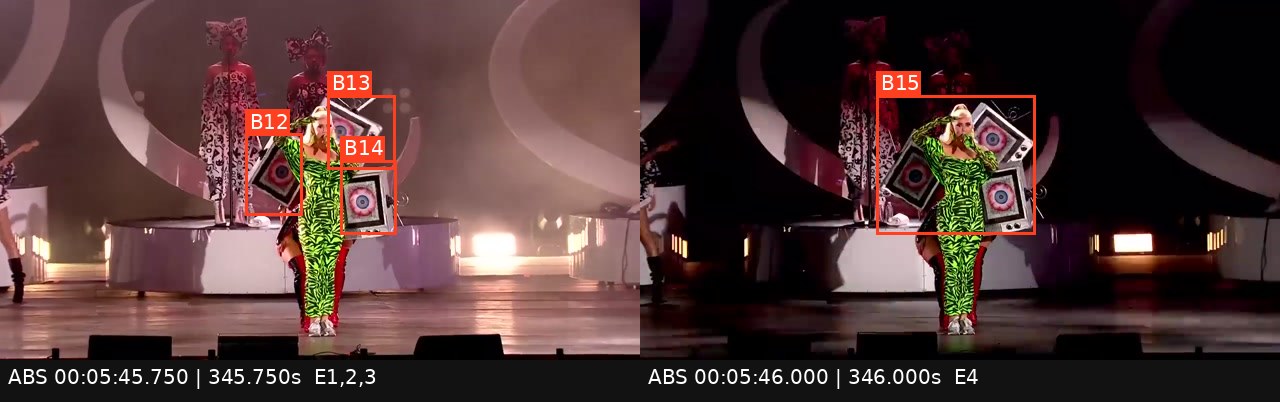}\\[1pt]
{\scriptsize\sffamily\color{cInkB} Evidence card of step~7 (05:45.75, 05:46): boxes B12--B14 mark one hood each, and B15 all three.}
\end{minipage}
\caption{\textbf{Complete trajectory: two \finding{}s that disagree on a count.} A one-minute, low-resolution Inspect reports four hoods; high-resolution Inspects of the seconds after 05:45, and a crop, report three. A second ``four'' counts the dancers earlier in the song, and tracking them into the last shot at 2 and 4\,fps settles the count.}
\label{fig:traj:1087}
\end{figure*}

\begin{figure*}[p]
\trhead{lvbench\_1439}{entity recognition, 33-min talk-show clip, annotated 24:30--24:50}%
  {What does the man put on the table for people to see?}%
  {(A) An empty box\quad (B) A box of needles\quad (C) A box of sweets\quad (D) A box of cards}%
  {Ground truth \textbf{B}.\quad \method~D\,\trno\quad Gemini Agentic~B\,\trok\quad Static~D\,\trno\quad VideoSeek~B\,\trok\hfill\textcolor{cInkB}{5 \planner turns, 69.9k \planner input tokens}}
\par\vspace{4pt}
\begin{minipage}[t]{0.625\textwidth}\footnotesize\raggedright
\begin{trsteps}
\trstep{1}\trtool{get\_transcript}() \trnote{at the annotated time: [1460.4--1467.3] ``Yeah, Jimmy, we brought some pins. There's pins. Could you grab these little pins right here?''}
\trstep{2}\trthink{I'm scrutinizing transcript lines mentioning ``box,'' especially around 350-360, where a deck is pulled out. I'm also looking at the needle and thread trick around 760-800, and the card-related actions near 1180 and 1320-1345. I'm trying to reconcile these with the question's options.}\\
  \trtool{overview}(query ``What does the man put on the table for people to see? Box of needles, empty box, box of sweets, box of cards'')\trread{1,024 frames + audio track: 67,584 video + 49,392 audio tokens}\\
  \trfind{}{``Throughout the video, David Blaine performs magic tricks on The Tonight Show. [\ldots] Blaine places a box of cards in the center of the table for everyone to observe (around 05:24). Later segments involve cards placed face up on the table, an empty tuck box discarded during an audience trick, and containers of pins and nails placed on the host desk.''}
  Candidates, in order: 05:15--05:45 ``[\ldots] places the box of cards onto the center of the black table [\ldots] so everyone can watch it before Jimmy names a card''; 03:55--04:30 (cards discarded on the table); 22:10--22:30 (an empty card box); 24:25--24:50 ``Jimmy Fallon pulls out a small clear plastic case containing straight pins and places it on the desk beside Blaine'', relevance ``Context for the `Box of needles' distractor, although the objects are straight pins rather than sewing needles and are brought out by Jimmy''; 29:05--29:45 (a box of nails).
\trstep{3}\trtool{inspect\_frames}(315--340\,s, 1\,fps, low; task ``Observe David Blaine placing an item/box on the table and what he says or does with it for people to see.'')\trread{25 frames + audio: 1,650 video + 625 audio tokens}\\
  \trfind{ok}{``David Blaine slides the deck of cards into its card box and places it flat on the table in plain view. [\ldots]''}
\trstep{4}\trthink{I'm currently examining a temporal window around the 1175 to 1205 mark. [\ldots]}\\
  \trtool{inspect\_frames}(1170--1195\,s, 1\,fps, low; task ``What does Jimmy open or what is being shown here?'')\trread{25 frames + audio: 1,650 video + 625 audio tokens}\\
  \trfind{ok}{``Jimmy Fallon unfolds a small folded piece of a playing card that was pulled from David Blaine's mouth and matches it to a torn playing card (Joker) [\ldots]''}
\trstep{5}Answer \textbf{D} \trno\enspace\trnote{the fourth candidate, which covers the annotated window, was never inspected}
\end{trsteps}
\end{minipage}\hfill
\begin{minipage}[t]{0.355\textwidth}\vspace{3pt}
\includegraphics[width=\linewidth]{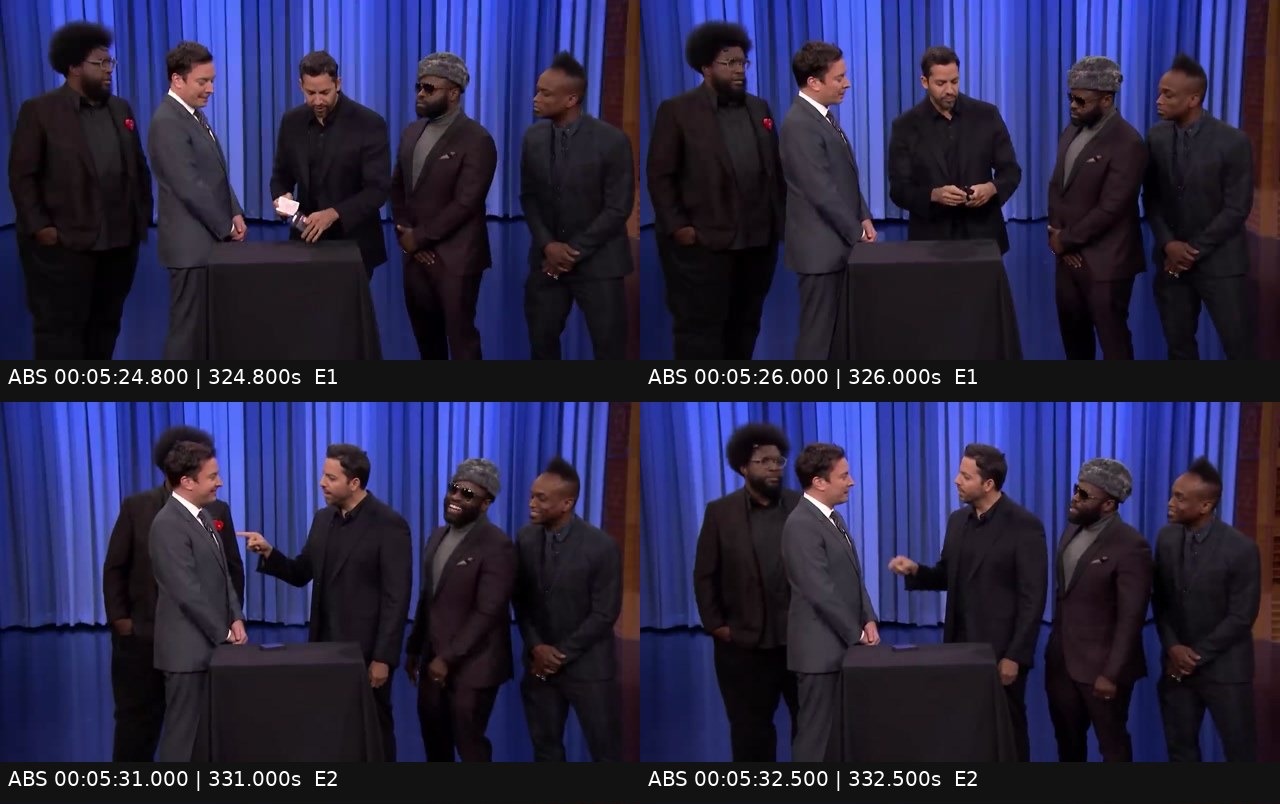}\\[1pt]
{\scriptsize\sffamily\color{cInkB} Four of the six frames on the evidence card of step~3 (05:24.8--05:32.5): the box of cards is placed on the table and stays there. Nothing from 24:25--24:50 reached the \planner except the Overview candidate text.}
\end{minipage}
\caption{\textbf{Complete trajectory: a typical failure.} The question admits two boxes. Overview lists both but ranks the box of cards first and describes the annotated one as context for a distractor, and the \planner inspects only the first candidate.}
\label{fig:traj:1439}
\end{figure*}

\Cref{fig:traj:1248,fig:traj:734,fig:traj:598,fig:traj:1087,fig:traj:1439} reproduce five complete trajectories from the full LVBench run reported in \cref{tab:lvbench}.
They are selected, not sampled: we chose them to show each kind of call doing its job (a count settled by one Overview and one Inspect, a \finding that contradicts the option the scene first suggests, several occurrences of an event told apart, two \finding{}s that disagree on a count and the denser look that settles it) and one typical error.
Three of the four correct answers are on questions at least two baselines miss, which is not typical (\method is right and Gemini Agentic wrong on 57 questions, the reverse on 40); \cref{fig:suppl:looked} shows questions drawn at random.
Calls, \planner thought summaries and \finding text are quoted verbatim from the trace and shortened only at [\ldots]; text in grey brackets is ours.
The \planner never receives the frames counted in teal; it receives the \finding and the evidence card shown beside each trajectory.

\paragraph{Counting (\cref{fig:traj:1248}).}
The vlog has no speech in the annotated window, so the \planner calls Overview, which locates the meal at 18:43--19:08 and the packing of leftovers right after it, but describes the meal as eating ``the sandwich''.
The \planner inspects one interval spanning the eating and the packing, and asks for both counts.
The \finding reports four quarters on the plate, two eaten and two packed, each claim tied to an interval, and the card shows the frames behind each claim.
Gemini Agentic and VideoSeek both answer (B), one piece.

\paragraph{A \finding that contradicts the expected option (\cref{fig:traj:734}).}
The film has no dialogue, and its transcript holds 23 fragments of noise, so the \planner starts with Overview.
Overview points to two scenes: a ladybug rescue from a spider web at about 44--47 minutes, and mantises hunting ladybugs at about 68--71.
The web scene looks like option (A), which is the answer all three baselines give.
The \planner inspects it, and two high-resolution \finding{}s report the sizes the other way round: the trapped ladybug is the smaller one, and the larger one rescues it.
The \planner notes the contradiction, runs Overview again with a new query, and inspects the mantis scene, where a smaller ladybug knocks a larger one out of the mantis's claws.
Two earlier \finding{}s on the web scene (steps 3 and 6) also see a larger ladybug caught on the web at the start of the scene; the two that look at the rescue itself agree on who saves whom, and the evidence cards bear them out.

\paragraph{Several occurrences, one asked for (\cref{fig:traj:598}).}
The video strings together auditions from several editions of a talent show, and nothing on screen numbers the contestants.
The transcript holds the lyrics of three songs before 4:20, and the \planner inspects the third: the sub-agent reads the caption, ``Creep'', but reports \texttt{uncertain} on whether this is the third contestant rather than asserting it.
The \planner then reads the first two captions with two 30-frame Inspects; their songs are the distractors (B) and (D), and the lyrics leave no room for a fourth song before Creep.
All four systems answer correctly; we show it for the \texttt{uncertain} status, which marks exactly the part of the task the interval could not settle.

\paragraph{Two \finding{}s that disagree on a count (\cref{fig:traj:1087}).}
The transcript places the song at 05:43, so the \planner skips Overview.
A one-minute, low-resolution Inspect reports four TV hoods after 05:45, Gemini Agentic's answer, although its own box at 05:46 holds three; a high-resolution Inspect at 2\,fps and a crop of that box report three.
A sparse pass over the preceding half minute then counts four hooded dancers, earlier in the song, and the \planner tracks them through 05:43--05:47 at 2 and 4\,fps: three hoods remain in the last shot.
Static answers (D).

\paragraph{A typical failure (\cref{fig:traj:1439}).}
The question admits two boxes: a box of cards placed on the table at 05:24, and a case of pins at about 24:30, which the annotation and option (B) point to.
The \planner's thought lists the transcript's mentions of a box, and the pins are never called one.
Overview ranks the box of cards first and lists the pins fourth, as ``context for the `Box of needles' distractor''; the \planner inspects only the first candidate and answers (D).
Gemini Agentic and VideoSeek answer (B), Static (D).
Of the 35 wrong answers on which no Inspect overlaps the annotated window, 24 called Overview, and in 12 of these an Overview candidate covered the window and was never inspected.

\subsection{Behaviour over the full run}
\label{sec:suppl:traj:stats}

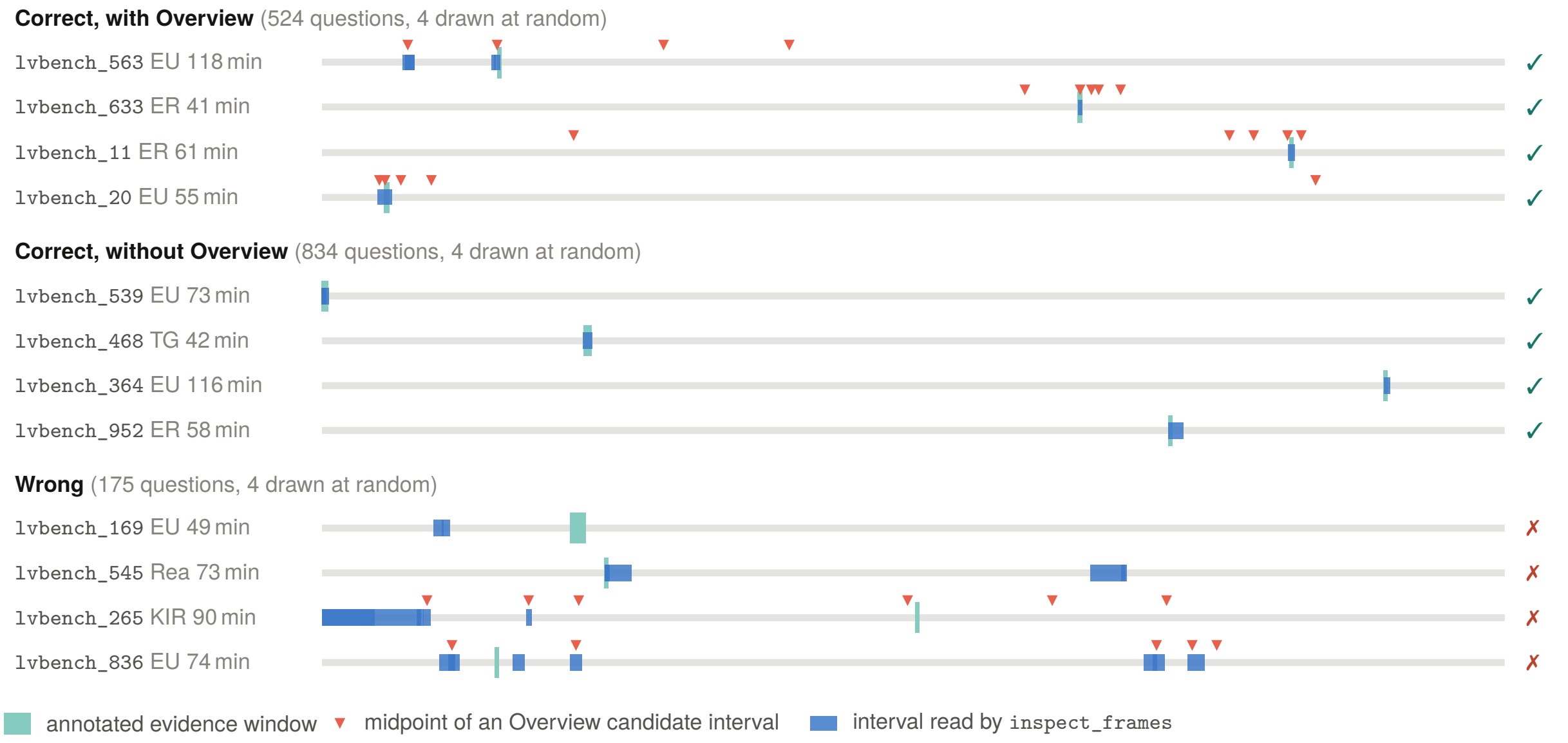
\begin{figure*}[htbp]
  \centering
  \begin{adjustbox}{max width=\linewidth}
\begin{tikzpicture}[x=1cm,y=1cm,font=\sffamily\scriptsize]
\node[anchor=west,font=\sffamily\bfseries\scriptsize,text=cInk] at (0,0.00) {Correct, with Overview \textcolor{cInkC}{\mdseries (524 questions, 4 drawn at random)}};
\node[anchor=west,text=cInkB] at (0,-0.47) {\texttt{lvbench\_563} \textcolor{cInkC}{EU 118\,min}};
\draw[cGrid,line width=2.2pt] (3.55,-0.47) -- (16.75,-0.47);
\fill[cTeal!55] (5.503,-0.65) rectangle (5.553,-0.30);
\fill[cOurs] (5.505,-0.34) -- ++(-0.06,0.12) -- ++(0.12,0) -- cycle;
\fill[cOurs] (4.508,-0.34) -- ++(-0.06,0.12) -- ++(0.12,0) -- cycle;
\fill[cOurs] (7.362,-0.34) -- ++(-0.06,0.12) -- ++(0.12,0) -- cycle;
\fill[cOurs] (8.765,-0.34) -- ++(-0.06,0.12) -- ++(0.12,0) -- cycle;
\fill[cPlan,opacity=.85] (5.472,-0.56) rectangle (5.537,-0.39);
\fill[cPlan,opacity=.85] (4.478,-0.56) rectangle (4.543,-0.39);
\fill[cPlan,opacity=.85] (4.478,-0.56) rectangle (4.543,-0.39);
\fill[cPlan,opacity=.85] (4.534,-0.56) rectangle (4.581,-0.39);
\fill[cPlan,opacity=.85] (4.451,-0.56) rectangle (4.506,-0.39);
\fill[cPlan,opacity=.85] (4.511,-0.56) rectangle (4.551,-0.39);
\fill[cPlan,opacity=.85] (4.542,-0.56) rectangle (4.582,-0.39);
\fill[cPlan,opacity=.85] (5.444,-0.56) rectangle (5.490,-0.39);
\node[anchor=west] at (16.87,-0.47) {\textcolor{cTealInk}{\ding{51}}};
\node[anchor=west,text=cInkB] at (0,-0.97) {\texttt{lvbench\_633} \textcolor{cInkC}{ER 41\,min}};
\draw[cGrid,line width=2.2pt] (3.55,-0.97) -- (16.75,-0.97);
\fill[cTeal!55] (11.981,-1.15) rectangle (12.035,-0.80);
\fill[cOurs] (11.394,-0.84) -- ++(-0.06,0.12) -- ++(0.12,0) -- cycle;
\fill[cOurs] (12.011,-0.84) -- ++(-0.06,0.12) -- ++(0.12,0) -- cycle;
\fill[cOurs] (12.137,-0.84) -- ++(-0.06,0.12) -- ++(0.12,0) -- cycle;
\fill[cOurs] (12.218,-0.84) -- ++(-0.06,0.12) -- ++(0.12,0) -- cycle;
\fill[cOurs] (12.463,-0.84) -- ++(-0.06,0.12) -- ++(0.12,0) -- cycle;
\fill[cPlan,opacity=.85] (11.986,-1.06) rectangle (12.035,-0.89);
\node[anchor=west] at (16.87,-0.97) {\textcolor{cTealInk}{\ding{51}}};
\node[anchor=west,text=cInkB] at (0,-1.48) {\texttt{lvbench\_11} \textcolor{cInkC}{ER 61\,min}};
\draw[cGrid,line width=2.2pt] (3.55,-1.48) -- (16.75,-1.48);
\fill[cTeal!55] (14.346,-1.65) rectangle (14.396,-1.31);
\fill[cOurs] (6.359,-1.35) -- ++(-0.06,0.12) -- ++(0.12,0) -- cycle;
\fill[cOurs] (13.678,-1.35) -- ++(-0.06,0.12) -- ++(0.12,0) -- cycle;
\fill[cOurs] (13.948,-1.35) -- ++(-0.06,0.12) -- ++(0.12,0) -- cycle;
\fill[cOurs] (14.326,-1.35) -- ++(-0.06,0.12) -- ++(0.12,0) -- cycle;
\fill[cOurs] (14.479,-1.35) -- ++(-0.06,0.12) -- ++(0.12,0) -- cycle;
\fill[cPlan,opacity=.85] (14.335,-1.57) rectangle (14.407,-1.39);
\node[anchor=west] at (16.87,-1.48) {\textcolor{cTealInk}{\ding{51}}};
\node[anchor=west,text=cInkB] at (0,-1.98) {\texttt{lvbench\_20} \textcolor{cInkC}{EU 55\,min}};
\draw[cGrid,line width=2.2pt] (3.55,-1.98) -- (16.75,-1.98);
\fill[cTeal!55] (4.239,-2.15) rectangle (4.303,-1.81);
\fill[cOurs] (4.191,-1.85) -- ++(-0.06,0.12) -- ++(0.12,0) -- cycle;
\fill[cOurs] (4.255,-1.85) -- ++(-0.06,0.12) -- ++(0.12,0) -- cycle;
\fill[cOurs] (4.431,-1.85) -- ++(-0.06,0.12) -- ++(0.12,0) -- cycle;
\fill[cOurs] (4.771,-1.85) -- ++(-0.06,0.12) -- ++(0.12,0) -- cycle;
\fill[cOurs] (14.639,-1.85) -- ++(-0.06,0.12) -- ++(0.12,0) -- cycle;
\fill[cPlan,opacity=.85] (4.171,-2.06) rectangle (4.331,-1.89);
\node[anchor=west] at (16.87,-1.98) {\textcolor{cTealInk}{\ding{51}}};
\node[anchor=west,font=\sffamily\bfseries\scriptsize,text=cInk] at (0,-2.60) {Correct, without Overview \textcolor{cInkC}{\mdseries (834 questions, 4 drawn at random)}};
\node[anchor=west,text=cInkB] at (0,-3.08) {\texttt{lvbench\_539} \textcolor{cInkC}{EU 73\,min}};
\draw[cGrid,line width=2.2pt] (3.55,-3.08) -- (16.75,-3.08);
\fill[cTeal!55] (3.550,-3.25) rectangle (3.622,-2.91);
\fill[cPlan,opacity=.85] (3.550,-3.17) rectangle (3.625,-2.99);
\fill[cPlan,opacity=.85] (3.550,-3.17) rectangle (3.595,-2.99);
\fill[cPlan,opacity=.85] (3.545,-3.17) rectangle (3.585,-2.99);
\fill[cPlan,opacity=.85] (3.548,-3.17) rectangle (3.588,-2.99);
\node[anchor=west] at (16.87,-3.08) {\textcolor{cTealInk}{\ding{51}}};
\node[anchor=west,text=cInkB] at (0,-3.58) {\texttt{lvbench\_468} \textcolor{cInkC}{TG 42\,min}};
\draw[cGrid,line width=2.2pt] (3.55,-3.58) -- (16.75,-3.58);
\fill[cTeal!55] (6.471,-3.75) rectangle (6.561,-3.41);
\fill[cPlan,opacity=.85] (6.465,-3.67) rectangle (6.566,-3.49);
\node[anchor=west] at (16.87,-3.58) {\textcolor{cTealInk}{\ding{51}}};
\node[anchor=west,text=cInkB] at (0,-4.08) {\texttt{lvbench\_364} \textcolor{cInkC}{EU 116\,min}};
\draw[cGrid,line width=2.2pt] (3.55,-4.08) -- (16.75,-4.08);
\fill[cTeal!55] (15.393,-4.25) rectangle (15.443,-3.91);
\fill[cPlan,opacity=.85] (15.401,-4.17) rectangle (15.467,-3.99);
\node[anchor=west] at (16.87,-4.08) {\textcolor{cTealInk}{\ding{51}}};
\node[anchor=west,text=cInkB] at (0,-4.58) {\texttt{lvbench\_952} \textcolor{cInkC}{ER 58\,min}};
\draw[cGrid,line width=2.2pt] (3.55,-4.58) -- (16.75,-4.58);
\fill[cTeal!55] (12.994,-4.75) rectangle (13.044,-4.41);
\fill[cPlan,opacity=.85] (12.995,-4.67) rectangle (13.166,-4.49);
\node[anchor=west] at (16.87,-4.58) {\textcolor{cTealInk}{\ding{51}}};
\node[anchor=west,font=\sffamily\bfseries\scriptsize,text=cInk] at (0,-5.20) {Wrong \textcolor{cInkC}{\mdseries (175 questions, 4 drawn at random)}};
\node[anchor=west,text=cInkB] at (0,-5.67) {\texttt{lvbench\_169} \textcolor{cInkC}{EU 49\,min}};
\draw[cGrid,line width=2.2pt] (3.55,-5.67) -- (16.75,-5.67);
\fill[cTeal!55] (6.315,-5.84) rectangle (6.493,-5.50);
\fill[cPlan,opacity=.85] (4.888,-5.76) rectangle (4.977,-5.58);
\fill[cPlan,opacity=.85] (4.799,-5.76) rectangle (4.910,-5.58);
\node[anchor=west] at (16.87,-5.67) {\textcolor{cOursInk}{\ding{55}}};
\node[anchor=west,text=cInkB] at (0,-6.17) {\texttt{lvbench\_545} \textcolor{cInkC}{Rea 73\,min}};
\draw[cGrid,line width=2.2pt] (3.55,-6.17) -- (16.75,-6.17);
\fill[cTeal!55] (6.701,-6.34) rectangle (6.751,-6.00);
\fill[cPlan,opacity=.85] (6.709,-6.26) rectangle (7.010,-6.08);
\fill[cPlan,opacity=.85] (6.709,-6.26) rectangle (6.764,-6.08);
\fill[cPlan,opacity=.85] (6.712,-6.26) rectangle (6.752,-6.08);
\fill[cPlan,opacity=.85] (12.126,-6.26) rectangle (12.517,-6.08);
\fill[cPlan,opacity=.85] (12.472,-6.26) rectangle (12.532,-6.08);
\node[anchor=west] at (16.87,-6.17) {\textcolor{cOursInk}{\ding{55}}};
\node[anchor=west,text=cInkB] at (0,-6.67) {\texttt{lvbench\_265} \textcolor{cInkC}{KIR 90\,min}};
\draw[cGrid,line width=2.2pt] (3.55,-6.67) -- (16.75,-6.67);
\fill[cTeal!55] (10.168,-6.84) rectangle (10.218,-6.50);
\fill[cOurs] (4.723,-6.54) -- ++(-0.06,0.12) -- ++(0.12,0) -- cycle;
\fill[cOurs] (5.857,-6.54) -- ++(-0.06,0.12) -- ++(0.12,0) -- cycle;
\fill[cOurs] (6.416,-6.54) -- ++(-0.06,0.12) -- ++(0.12,0) -- cycle;
\fill[cOurs] (10.086,-6.54) -- ++(-0.06,0.12) -- ++(0.12,0) -- cycle;
\fill[cOurs] (11.700,-6.54) -- ++(-0.06,0.12) -- ++(0.12,0) -- cycle;
\fill[cOurs] (12.976,-6.54) -- ++(-0.06,0.12) -- ++(0.12,0) -- cycle;
\fill[cPlan,opacity=.85] (5.832,-6.76) rectangle (5.894,-6.58);
\fill[cPlan,opacity=.85] (4.679,-6.76) rectangle (4.765,-6.58);
\fill[cPlan,opacity=.85] (3.550,-6.76) rectangle (4.139,-6.58);
\fill[cPlan,opacity=.85] (4.139,-6.76) rectangle (4.679,-6.58);
\fill[cPlan,opacity=.85] (4.616,-6.76) rectangle (4.656,-6.58);
\fill[cPlan,opacity=.85] (4.029,-6.76) rectangle (4.139,-6.58);
\fill[cPlan,opacity=.85] (3.550,-6.76) rectangle (4.029,-6.58);
\fill[cPlan,opacity=.85] (4.610,-6.76) rectangle (4.650,-6.58);
\node[anchor=west] at (16.87,-6.67) {\textcolor{cOursInk}{\ding{55}}};
\node[anchor=west,text=cInkB] at (0,-7.17) {\texttt{lvbench\_836} \textcolor{cInkC}{EU 74\,min}};
\draw[cGrid,line width=2.2pt] (3.55,-7.17) -- (16.75,-7.17);
\fill[cTeal!55] (5.478,-7.34) rectangle (5.528,-7.00);
\fill[cOurs] (5.001,-7.04) -- ++(-0.06,0.12) -- ++(0.12,0) -- cycle;
\fill[cOurs] (6.385,-7.04) -- ++(-0.06,0.12) -- ++(0.12,0) -- cycle;
\fill[cOurs] (12.864,-7.04) -- ++(-0.06,0.12) -- ++(0.12,0) -- cycle;
\fill[cOurs] (13.262,-7.04) -- ++(-0.06,0.12) -- ++(0.12,0) -- cycle;
\fill[cOurs] (13.536,-7.04) -- ++(-0.06,0.12) -- ++(0.12,0) -- cycle;
\fill[cPlan,opacity=.85] (5.678,-7.26) rectangle (5.812,-7.08);
\fill[cPlan,opacity=.85] (6.318,-7.26) rectangle (6.452,-7.08);
\fill[cPlan,opacity=.85] (4.964,-7.26) rectangle (5.038,-7.08);
\fill[cPlan,opacity=.85] (4.860,-7.26) rectangle (4.994,-7.08);
\fill[cPlan,opacity=.85] (12.822,-7.26) rectangle (12.956,-7.08);
\fill[cPlan,opacity=.85] (13.209,-7.26) rectangle (13.402,-7.08);
\fill[cPlan,opacity=.85] (5.008,-7.26) rectangle (5.083,-7.08);
\fill[cPlan,opacity=.85] (4.986,-7.26) rectangle (5.026,-7.08);
\fill[cPlan,opacity=.85] (12.718,-7.26) rectangle (12.866,-7.08);
\fill[cPlan,opacity=.85] (4.987,-7.26) rectangle (5.027,-7.08);
\node[anchor=west] at (16.87,-7.17) {\textcolor{cOursInk}{\ding{55}}};
\fill[cTeal!55] (0.0,-7.97) rectangle (0.30,-7.73);\node[anchor=west,text=cInkB] at (0.35,-7.85) {annotated evidence window};
\fill[cOurs] (3.75,-7.91) -- ++(-0.06,0.12) -- ++(0.12,0) -- cycle;\node[anchor=west,text=cInkB] at (3.9,-7.85) {midpoint of an Overview candidate interval};
\fill[cPlan,opacity=.85] (9.0,-7.93) rectangle (9.3,-7.77);\node[anchor=west,text=cInkB] at (9.35,-7.85) {interval read by \texttt{inspect\_frames}};
\end{tikzpicture}
\end{adjustbox}
  \caption{\textbf{Where the \planner looked.} Twelve LVBench questions, four drawn at random (fixed seed) from each group among the questions with an annotated evidence window; each track spans its whole video. Task abbreviations: ER entity recognition, EU event understanding, KIR key information retrieval, Rea reasoning, TG temporal grounding.}
  \label{fig:suppl:looked}
\end{figure*}

\begin{figure*}[htbp]
  \centering
  \begin{adjustbox}{max width=\linewidth}
\begin{tikzpicture}
\begin{axis}[name=a, font=\sffamily\scriptsize, axis line style={cInkC}, tick style={cInkC}, tick label style={font=\sffamily\scriptsize, text=cInkB}, label style={text=cInkB}, title style={font=\sffamily\scriptsize, text=cInk}, width=5.7cm, height=3.9cm, ybar, bar width=3.2pt, ymin=0, ymax=420,
  xlabel={\planner turns per question}, ylabel={questions}, xmin=1, xmax=23,
  xtick={2,6,10,14,18,22}, xticklabels={2,6,10,14,18,22}, ytick={0,100,200,300,400}, yticklabels={0,100,200,300,400}, title={\textbf{(a)} \planner turns}]
\addplot[fill=cPlan,draw=none] coordinates {(2,143) (3,374) (4,325) (5,162) (6,119) (7,71) (8,41) (9,42) (10,31) (11,26) (12,21) (13,16) (14,24) (15,23) (16,32) (17,34) (18,26) (19,31) (20,3) (21,2) (22,3)};
\draw[cOursInk,dashed,line width=.6pt] (axis cs:6.12,0) -- (axis cs:6.12,420);
\node[anchor=north west,font=\sffamily\tiny,text=cOursInk] at (axis cs:6.42,410) {mean 6.1};
\end{axis}
\begin{axis}[name=b, at={(a.east)}, anchor=west, xshift=0.9cm, font=\sffamily\scriptsize, axis line style={cInkC}, tick style={cInkC}, tick label style={font=\sffamily\scriptsize, text=cInkB}, label style={text=cInkB}, title style={font=\sffamily\scriptsize, text=cInk}, width=5.7cm, height=3.9cm, ybar,
  bar width=3.2pt, ymin=0, ymax=460, xlabel={perception calls per question}, xmin=-1, xmax=16,
  xtick={0,3,6,9,12,15}, xticklabels={0,3,6,9,12,15+}, ytick={0,200,400}, yticklabels={0,200,400},
  title={\textbf{(b)} Perception calls}]
\addplot[fill=cTeal,draw=none] coordinates {(0,23) (1,419) (2,386) (3,208) (4,125) (5,72) (6,53) (7,36) (8,34) (9,33) (10,28) (11,39) (12,60) (13,20) (14,10) (15,3)};
\draw[cOursInk,dashed,line width=.6pt] (axis cs:3.67,0) -- (axis cs:3.67,460);
\node[anchor=north west,font=\sffamily\tiny,text=cOursInk] at (axis cs:3.92,450) {mean 3.7};
\end{axis}
\begin{axis}[at={(b.east)}, anchor=west, xshift=1.75cm, font=\sffamily\scriptsize, axis line style={cInkC}, tick style={cInkC}, tick label style={font=\sffamily\scriptsize, text=cInkB}, label style={text=cInkB}, title style={font=\sffamily\scriptsize, text=cInk}, width=4.9cm, height=3.9cm, xbar stacked,
  bar width=7pt, xmin=0, xmax=100, ytick={0,1,2}, yticklabels={{\texttt{ok} (4{,}123)},{\texttt{absent} (560)},{\texttt{uncertain} (141)}}, y dir=reverse, ymin=-0.6, ymax=2.6,
  xtick={0,25,50,75,100}, xticklabels={0,25,50,75,100}, xlabel={\% of Findings with that status},
  title={\textbf{(c)} \planner's next step after a \finding},
  legend style={font=\sffamily\tiny, at={(0.42,-0.40)}, anchor=north, legend columns=5, draw=none,
  /tikz/every even column/.append style={column sep=2.5pt}}, legend cell align=left]
\addplot[fill=cSlate,draw=white,line width=.3pt] coordinates {(31.12,0) (13.93,1) (15.60,2)};\addlegendentry{answer}
\addplot[fill=cPlan,draw=white,line width=.3pt] coordinates {(56.59,0) (68.39,1) (60.99,2)};\addlegendentry{Inspect}
\addplot[fill=cPlan!40,draw=white,line width=.3pt] coordinates {(9.36,0) (9.82,1) (17.02,2)};\addlegendentry{crop}
\addplot[fill=cOurs,draw=white,line width=.3pt] coordinates {(1.79,0) (7.14,1) (2.13,2)};\addlegendentry{Overview}
\addplot[fill=cAmber,draw=white,line width=.3pt] coordinates {(1.14,0) (0.71,1) (4.26,2)};\addlegendentry{transcript}

\end{axis}
\end{tikzpicture}
\end{adjustbox}
  \caption{\textbf{\planner behaviour over all 1,549 LVBench questions.} (a) \planner turns, including the answer. (b) Perception calls (Overview, Inspect and crop). (c) For each \finding status, the \planner's next call, or ``answer'' when the \finding was the last observation before answering.}
  \label{fig:suppl:behaviour}
\end{figure*}
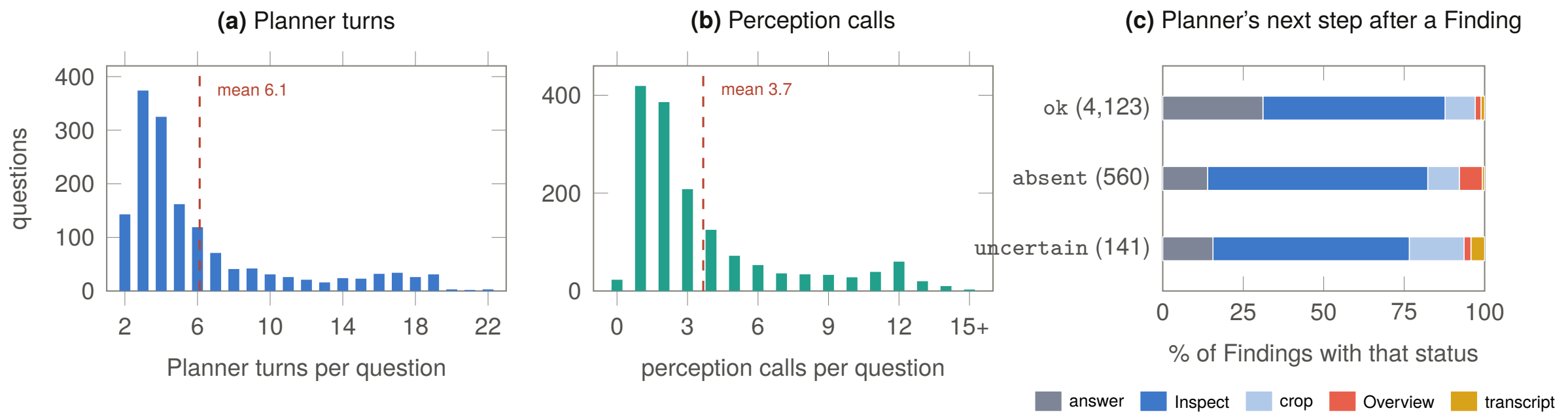

\Cref{fig:suppl:looked} shows where the \planner spent its perception on twelve questions drawn at random.
On the eight correct ones, Inspect intervals sit on or next to the annotated window; where Overview was called, its candidates spread across the video and the \planner inspects one or two of them.
The four errors show the two ways the loop fails: three never inspect the annotated window, and one inspects it and still answers wrongly (\texttt{lvbench\_545}, whose first Inspect covers the window).
Over the run, at least one Inspect interval overlaps the annotated window on 1,251 questions, of which 89.0\% are answered correctly; on the 255 questions where every Inspect misses it, 86.3\% are, so a missed window rarely means a failed question: the evidence a question needs can also appear outside the one window the benchmark annotates.

In \cref{fig:suppl:behaviour}, the most common counts are three and four turns, and the mean is 6.1; 23 questions are answered from the transcript alone, 21 of them correctly.
After an \texttt{ok} \finding the \planner answers immediately 31\% of the time, after an \texttt{absent} or \texttt{uncertain} one 14\% and 16\%; after 383 of the 560 \texttt{absent} \finding{}s it inspects again.
Overview is called on 632 questions; 83.5\% of them are answered correctly against 92.0\% of the others, because the \planner calls Overview when the transcript does not locate the evidence, which also makes a question harder.
Neutral local tasks are asked for but not enforced: 441 of the 4,992 Inspect and crop tasks (8.8\%) name two or more options as whole words, as step~7 of \cref{fig:traj:1087} does with counts from earlier \finding{}s.

\clearpage

\section{Component analysis}
\label{sec:exp:ablation}

\begin{table}[htbp]
  \centering
  \small
  \setlength{\tabcolsep}{3pt}
  \begin{tabular*}{\linewidth}{@{\extracolsep{\fill}}llc@{}}
    \toprule
    Component & Variant & Acc. \\
    \midrule
    Overview candidates & forced / optional / none & \textbf{80.0} / 77.0 / 76.0 \\
    Overview content$^\dagger$ & timeline + cand. / cand.\ only & \textbf{81.0} / 77.0 \\
    Tool interface$^\dagger$ & atomic / batched (2 runs) & \textbf{80.5} / 75.0 \\
    Inspect resolution & high / medium / low & 80.0 / 80.0 / \textbf{81.5} \\
    Inspect sampling & 2\,fps / 1\,fps & \textbf{81.5} / 81.0 \\
    Inspect audio + ASR & without / with & 80.0 / \textbf{82.5} \\
    \planner system prompt & host default / slim Skill & \textbf{81.5} / 80.0 \\
    Evidence-closure rule & without / with & 79.5 / \textbf{80.0} \\
    \bottomrule
  \end{tabular*}
  \caption{\textbf{Component ablations} with \method v0 and Gemini~3.0 Flash \dev{on dev-200} ($^\dagger$\dev{dev-100}). Low Inspect resolution uses 72.9\% fewer image tokens than high; the slim Skill prompt cuts \planner input by 46.8\% for 1.5 points; the evidence-closure rule cuts turns by 26.5\% and cost by 23.7\%. Repeating one configuration twice moved it by 3 questions (79 and 82 of 100), which is the resolution of this table.}
  \label{tab:ablation}
\end{table}

\Cref{tab:ablation} summarizes design choices made during development with an earlier version of the harness.
Which visual information should reach the \planner was measured on the same subset and is reported in \cref{tab:memory}.
Forcing Overview to propose at least one candidate interval is better than letting it return none, and keeping the global timeline in addition to candidates is better than candidates alone.
Separate atomic tools outperform a single batched inspection tool by 5.5 points.
Low-resolution Inspect is as accurate as high resolution while using 72.9\% fewer image tokens.
Giving Inspect the audio track and ASR improves accuracy by 2.5 points.
We also tested several \emph{prompt-level policies} that tell the \planner how to sample (density hints, joint multi-window inspection, mandatory cropping, explicit evidence-state receipts, neutral-task rules in the tool schema, dedicated counting tools).
None provided convincing evidence of improvement on this development subset, and several reduced accuracy by 3--5 points; we report them as a group because no individual policy separated from the baseline.
These development results motivated a minimal policy that leaves observation choices to the \planner, without establishing that this policy is generally optimal.
Raising the turn limit from 10 to 14 was measured separately on 20 questions repeated three times and moved the mean from 10.00 to 10.67 correct while increasing tool calls by 32\%, so the limit was left at the lower value.

\section{When does an agent help?}
\label{sec:exp:when}

Agentic perception is not free.
On LongVideoBench with Gemini~3.0, the earlier harness loses 2.1 points on 15-second clips and gains 1--3 points on videos longer than one minute.
On the long split of Video-MME, where the static model already reaches 85.2\%, it loses 4.0 points.
Questions that depend on global temporal order (for example, ordering of scenes) are hurt by local inspection, whereas questions that require finding and identifying a specific entity or event benefit most.
These results suggest routing between static ingestion and the agent by video length and question type, which we leave to future work.
This block was measured on an earlier version of the harness than the one in the main paper, so we report it for the direction of the effect rather than its magnitude.

\section{Inside Gemini's agentic mode}
\label{sec:suppl:agentic}
We analyzed traces of Gemini's agentic media-processing mode on ten LVBench questions with Gemini~3.7 Flash, one of the documented models for this mode.
The responses expose 46 thought signatures, of which only the ten final ones are decodable, and 28 thought summaries.
On average the model performs 1.8 internal navigation steps per question.
Tool-use token accounting reports image and text tokens but no audio tokens, and for 1,311 of the 1,429 full-LVBench questions where the mode navigated at all (92\%) the image-token count is an exact multiple of 66, the per-frame cost of low-resolution video; the remaining counts exceed a multiple of 66 by a small amount, so some navigated images are not plain low-resolution frames.
These observations suggest a server-side media index that offers timestamped transcripts and random access to frames, with the navigated frames returned into the reasoning model's context.

\begin{figure}[htbp]
  \centering
  \begin{adjustbox}{max width=\linewidth}
\begin{tikzpicture}
\begin{axis}[name=a, font=\sffamily\scriptsize, axis line style={cInkC}, tick style={cInkC}, tick label style={font=\sffamily\scriptsize, text=cInkB}, label style={text=cInkB}, title style={font=\sffamily\scriptsize, text=cInk}, ybar, bar width=10pt, ymin=0, xtick={0,1,2}, xticklabels={{with\\speech\\\textcolor{cInkC}{(1{,}077)}},{no\\speech\\\textcolor{cInkC}{(284)}},{no\\audio\\\textcolor{cInkC}{(175)}}}, xticklabel style={align=center, font=\sffamily\tiny}, xmin=-0.6, xmax=2.6, width=3.95cm, height=4.0cm, ymax=340, ytick={0,100,200,300}, yticklabels={0,100,200,300},
  ylabel={median frames read}, title={\textbf{(a)} Gemini Agentic}]
\addplot[fill=cAmber,draw=none] coordinates {(0,93.0) (1,285.5) (2,175.0)};
\node[above,font=\sffamily\tiny,text=cInkB] at (axis cs:0,93.0) {93};
\node[above,font=\sffamily\tiny,text=cInkB] at (axis cs:1,285.5) {286};
\node[above,font=\sffamily\tiny,text=cInkB] at (axis cs:2,175.0) {175};
\end{axis}
\begin{axis}[at={(a.east)}, anchor=west, xshift=1.2cm, font=\sffamily\scriptsize, axis line style={cInkC}, tick style={cInkC}, tick label style={font=\sffamily\scriptsize, text=cInkB}, label style={text=cInkB}, title style={font=\sffamily\scriptsize, text=cInk}, ybar, bar width=10pt, ymin=0, xtick={0,1,2}, xticklabels={{with\\speech\\\textcolor{cInkC}{(1{,}077)}},{no\\speech\\\textcolor{cInkC}{(284)}},{no\\audio\\\textcolor{cInkC}{(175)}}}, xticklabel style={align=center, font=\sffamily\tiny}, xmin=-0.6, xmax=2.6, width=3.95cm, height=4.0cm, ymax=100, ytick={0,25,50,75,100}, yticklabels={0,25,50,75,100},
  ylabel={\% of questions}, title={\textbf{(b)} \method: Overview called}]
\addplot[fill=cOurs,draw=none] coordinates {(0,28.8) (1,62.3) (2,81.1)};
\node[above,font=\sffamily\tiny,text=cInkB] at (axis cs:0,28.8) {29\%};
\node[above,font=\sffamily\tiny,text=cInkB] at (axis cs:1,62.3) {62\%};
\node[above,font=\sffamily\tiny,text=cInkB] at (axis cs:2,81.1) {81\%};
\end{axis}
\end{tikzpicture}
\end{adjustbox}
  \caption{\textbf{Two ways to navigate without a transcript.} LVBench questions split by whether our transcript has speech in the annotated evidence window, and the questions about videos with no audio track. (a) Frames Gemini Agentic reads into its context (image tool tokens / 66). (b) Share of questions on which \method's \planner calls Overview.}
  \label{fig:suppl:agentic}
\end{figure}
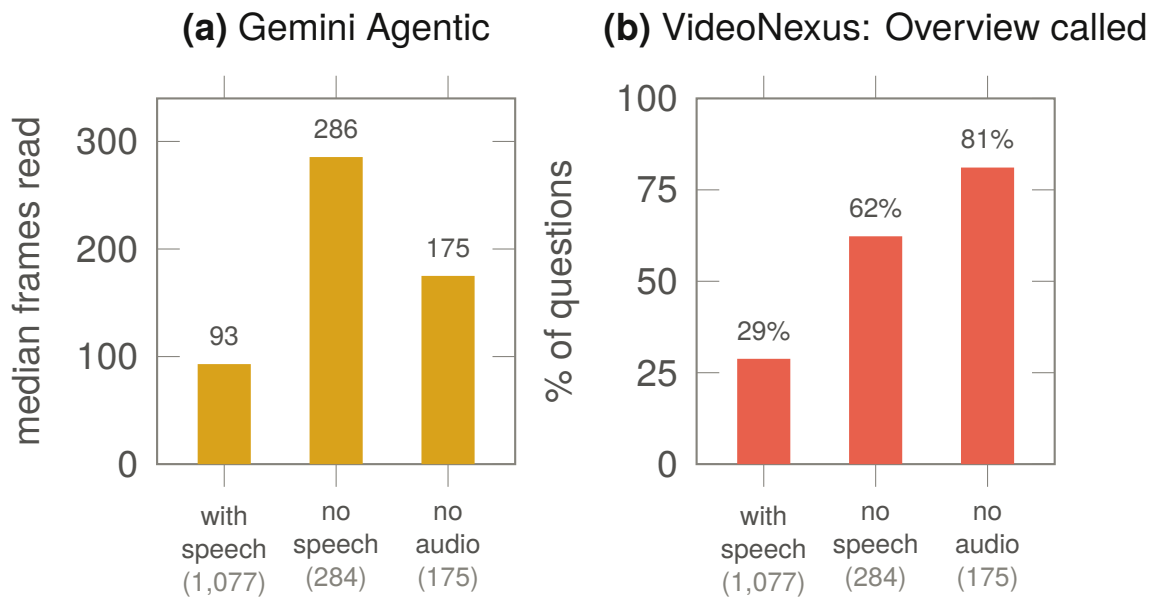

Over the full LVBench run its navigation is transcript-first.
1,353 of the 1,549 thought summaries refer to the transcript, ASR or subtitles; the tool-use text tokens, which carry what it retrieves from the transcript, have a median of 16.6k per question; and it reads a median of 93 frames on questions whose annotated evidence window our transcript places speech in, against 286 on questions whose window has no speech.
A visual scan is available, and 447 thought summaries describe sparse or low-frame-rate scanning. Transcript-based navigation is frequent, but the traces do not establish why the model selects this strategy.
\method's Overview makes the global scan cheap for the \planner, and the \planner uses it where the transcript cannot help: it fires on 29\% of the questions whose evidence window contains speech, on 62\% of those whose window does not, and on 81\% of the questions about the 12 videos without an audio track (\cref{fig:suppl:agentic}).

Its cost is dominated by reasoning tokens: on LVBench it averages 72.4k reasoning tokens per question, against 12.4k output tokens per question in total for \method, of which 2.6k are emitted by the \planner and the rest by short-lived perception calls.

\section{The native timeline probe}
\label{sec:suppl:timeline}

\method's \finding{}s are only composable if every perception call reports time on the same clock, so we
checked what the perception model does when it is given an interval of a longer video rather than a
clip. Across 12 probe responses with neutral prompts -- no instruction about which origin to use --
all 12 reported times on the absolute source timeline, including the 6 that were asked to state
their reference explicitly. The 6 responses given a physically cropped clip instead reported times
relative to that clip's own start. We therefore pass intervals by timestamp and never hand the model
a pre-cut file, which removes the offset bookkeeping that clip-based harnesses need.

\section{Limitations}
Each Inspect adds a model request and can increase latency; the available measurements do not establish a matched wall-clock comparison with static ingestion. We do not report a wall-clock comparison against the baseline: in our own runs only 22 of 160 median seconds per question are recorded API time, the rest being local decode, transcode and upload, and the baseline's per-call latency varied by 50\% across hosts with load, so neither number isolates the harness.

\paragraph{The benchmark has a text prior.}
A blind control given the question alone, with no video, audio or transcript, answers 40.5\% of full LVBench (\cref{tab:lvbench}). This demonstrates substantial performance without video. The control uses the same high thinking level as the main arms; its accuracy neither bounds all possible text-only strategies nor provides a causal decomposition of multimodal performance.
The VideoSeek re-run answered 52 of the 1{,}549 questions without a perception call and was correct on 25 (48\%). This self-selected subset further illustrates that some answers can be produced without viewing the video.

\paragraph{Costs are provider-specific.}
All dollar figures are list prices of one model; the image-tier confound and its bound are in \cref{sec:exp:transport}.

\paragraph{The baseline comparison is asymmetric on speech, in our favour.}
\method reads the native audio track and an ASR transcript; the VideoSeek re-run, following its released configuration, reads neither, and its own subtitle path filters windows with a disjunction where a conjunction is intended, so enabling it would place the entire transcript in every call. We therefore report the no-speech slices separately (\cref{sec:suppl:density}), where our margin is 3.9 points rather than the 8.7 it reaches where speech falls inside the evidence window -- this variation cannot quantify the modality advantage because the question groups differ and speech elsewhere may still aid localization. 12 of LVBench's 103 videos, covering 175 questions, carry no audio stream at all, so for those questions the asymmetry does not exist.

\paragraph{The ablations are paired but not held out.}
The ablations of \cref{sec:exp:coupling,sec:exp:density,sec:suppl:memory} are measured on the 200 dev questions used during development. Pairing controls question composition but does not remove development selection or configuration differences. These are not held-out results. We report discordant-pair counts where per-question records are available; the Gemini~3.8 card comparison has aggregate counts only.
The resumed baseline run also reveals answer variability: 55 of its 1{,}549 questions were answered twice at identical settings, and 7 of the 55 came back with a different answer, which cautions against overinterpreting small differences but does not estimate the variance of full-run accuracy. Where a duplicate exists we count the earliest-finished attempt, a rule fixed independently of whether it was correct.
Errors made inside a perception call persist unless the \planner decides to look again; overconfident \texttt{absent} findings and inconsistent counts across calls are the main residual errors.
All components are accessed through commercial APIs whose versions may change.
Development decisions were made on a 200-question subset of LVBench, which we report separately from full-set results.
On short clips and on questions that depend on the global order of events, a single static pass remains competitive or better.

\end{document}